\documentclass[10pt,journal,compsoc]{IEEEtran}

\ifCLASSOPTIONcompsoc
  \usepackage[nocompress]{cite}
\else
  \usepackage{cite}
\fi
\ifCLASSINFOpdf
  \usepackage[pdftex]{graphicx}
\else
  \usepackage[dvips]{graphicx}
\fi
\usepackage{amsmath}
\usepackage{array}
\usepackage{url}
\newcommand\MYhyperrefoptions{bookmarks=true,bookmarksnumbered=true,
pdfpagemode={UseOutlines},plainpages=false,pdfpagelabels=true,
colorlinks=true,linkcolor={black},citecolor={black},urlcolor={black},
pdftitle={Bare Demo of IEEEtran.cls for Computer Society Journals},
pdfsubject={Typesetting},
pdfauthor={Michael D. Shell},
pdfkeywords={Computer Society, IEEEtran, journal, LaTeX, paper,
             template}}
\ifCLASSINFOpdf
\usepackage[\MYhyperrefoptions,pdftex]{hyperref}
\else
\usepackage[\MYhyperrefoptions,breaklinks=true,dvips]{hyperref}
\usepackage{breakurl}
\fi

\usepackage{amsfonts}
\usepackage{textcomp}
\usepackage{stfloats}
\usepackage{verbatim}

\usepackage{algorithm, algpseudocode}
\usepackage{amssymb}
\usepackage{amsthm}
\usepackage{booktabs}
\usepackage{multirow}
\usepackage{subcaption}
\usepackage{float}
\usepackage{setspace}
\usepackage{framed}
\usepackage{bm}
\usepackage{color}

\usepackage{xurl}
\usepackage[capitalize]{cleveref}

\def\eg{\emph{e.g}.} 
\def\ie{\emph{i.e}.}

\def\etal{\emph{et al}.}

\begin{document}
%
\title{Adaptive Perturbation for Adversarial Attack}
%
%
%
%

\author{Zheng Yuan,~\IEEEmembership{Student Member,~IEEE,}
        Jie Zhang$^*$,~\IEEEmembership{Member,~IEEE,}
        Zhaoyan Jiang, Liangliang Li,
        Shiguang Shan,~\IEEEmembership{Fellow,~IEEE}
\IEEEcompsocitemizethanks{
\IEEEcompsocthanksitem J. Zhang is the corresponding author. \protect\\
\IEEEcompsocthanksitem Z. Yuan, J. Zhang and S. Shan are with Key Lab of Intelligent Information Processing of Chinese Academy of Sciences (CAS), Institute of Computing Technology, CAS, Beijing, 100190, China and University of China Academy of Sciences, Beijing, 100049, China. E-mail: zheng.yuan@vipl.ict.ac.cn, \{zhangjie, sgshan\}@ict.ac.cn. \protect\\
\IEEEcompsocthanksitem Z. Jiang and L. Li are with Tencent, Shenzhen, 518057, China. E-mail: \{zhaoyanjiang, apexli\}@tencent.com.}
\thanks{Manuscript received April 19, 2005; revised August 26, 2015.}}

%
%

\markboth{Journal of \LaTeX\ Class Files,~Vol.~14, No.~8, August~2015}%
{Shell \MakeLowercase{\textit{et al.}}: Bare Advanced Demo of IEEEtran.cls for IEEE Computer Society Journals}
%



\IEEEtitleabstractindextext{%
\begin{abstract}
  In recent years, the security of deep learning models achieves more and more attentions with the rapid development of neural networks, which are vulnerable to adversarial examples. Almost all existing gradient-based attack methods use the sign function in the generation to meet the requirement of perturbation budget on $L_\infty$ norm. However, we find that the sign function may be improper for generating adversarial examples since it modifies the exact gradient direction. Instead of using the sign function, we propose to directly utilize the exact gradient direction with a scaling factor for generating adversarial perturbations, which improves the attack success rates of adversarial examples even with fewer perturbations. At the same time, we also theoretically prove that this method can achieve better black-box transferability. Moreover, considering that the best scaling factor varies across different images, we propose an adaptive scaling factor generator to seek an appropriate scaling factor for each image, which avoids the computational cost for manually searching the scaling factor. Our method can be integrated with almost all existing gradient-based attack methods to further improve their attack success rates. Extensive experiments on the CIFAR10 and ImageNet datasets show that our method exhibits higher transferability and outperforms the state-of-the-art methods.
\end{abstract}

}

\maketitle

\IEEEdisplaynontitleabstractindextext

%
\IEEEpeerreviewmaketitle

\ifCLASSOPTIONcompsoc
\IEEEraisesectionheading{\section{Introduction}\label{sec:intro}}
\else
\section{Introduction}
\label{sec:intro}
\fi

%
%
%
%
\IEEEPARstart{W}{ith} the rapid progress and significant success in the deep learning in recent years, its security issue has attracted more and more attention. One of the most concerned security problems is its vulnerability to small, human-imperceptible adversarial noise~\cite{szegedy2014intriguing}, which implies severe risk of being attacked intentionally especially for technologies like face recognition and automatic driving. While it is important to study how to strengthen deep learning models to defend adversarial attack, it is equally important to explore how to attack these models. 

  Existing attack methods generate an adversarial example by adding to input some elaborately designed adversarial perturbations, which are usually generated either by a generative network~\cite{xiao2018generating, zhao2018generating, song2018constructing, ameya2019semantic, qiu2020semantivadv, xiao2021improving} or by the gradient-based optimization~\cite{goodfellow2015explaining, kurakin2017adversarial, madry2018towards, dong2018boosting, xie2019improving, dong2019evading, lin2020nesterov}. The latter, \ie, gradient-based methods, are the mainstream. Their key idea is generating the perturbation by exploiting the gradient computed via maximizing the loss function of the target task.

\begin{figure}[t]
    \centering
    \includegraphics[width=\columnwidth]{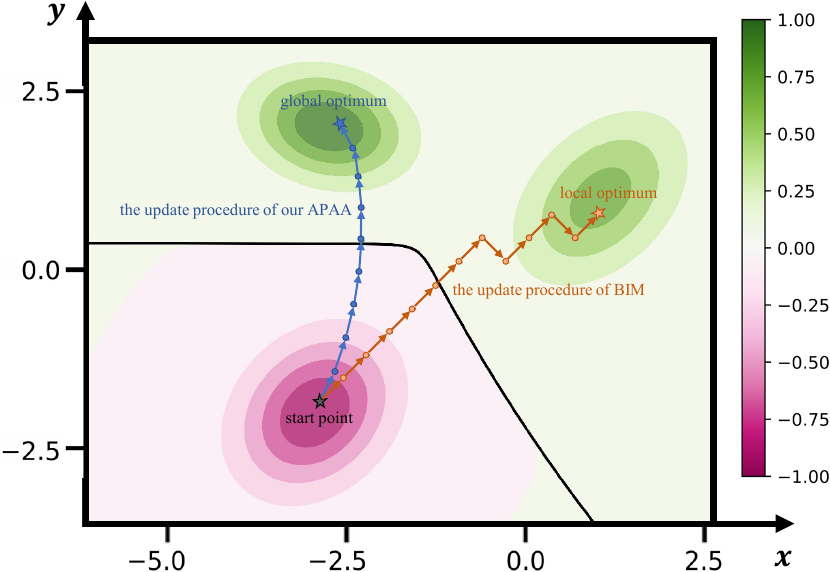}
    \caption{A two-dimensional toy example to illustrate the difference between our proposed APAA and existing sign-based methods, \eg, BIM~\cite{kurakin2017adversarial}. The loss function is composed of a mixture of Gaussian distributions, as described in~\cref{eqn:example}. The orange path and blue path represent the update process of BIM and our APAA when generating adversarial examples, respectively. The background color represents the contour of the loss function. During the adversarial attack, we aim to achieve an adversarial example with a larger loss value. Due to the limitation of the sign function, there are only eight possible update directions in the case of a two-dimensional space ((0, 1), (0, -1), (1, 1), (1, -1), (1, 0), (-1, -1), (-1, 0), (-1, 1)). The update direction of BIM is limited and not accurate enough, resulting in only reaching the sub-optimal end-point. Our method can not only obtain a more accurate update direction, but also adjust the step size adaptively. As a result, our APAA may reach the global optimum with a larger probability in fewer steps.}
    \label{fig:motivation}
  \end{figure}

  In these methods, since the gradient varies in different pixels, the sign function is usually used to normalize the gradient, which is convenient to set the step size of each step during the attack. Under the most commonly used $L_\infty$ norm setting in the adversarial attack, \ie, constraining the maximal $L_\infty$ norm of the generated adversarial perturbations, the use of the sign function can also leverage the largest perturbation budget to enhance the aggressiveness of adversarial examples.
  Due to the influence of the sign function, all values of the gradient are normalized into $\{0, +1, -1\}$. Although this approach can scale the gradient value, so as to make full use of the perturbation budget in the $L_\infty$ attack, the resultant update directions in adversarial attack are limited, there are only eight possible update directions in the case of a two-dimensional space. The inaccurate update direction may cause the generated adversarial examples to be sub-optimal (as shown in \cref{fig:motivation}).

  To solve the above problem of existing methods, we propose a method called Adaptive Perturbation for Adversarial Attack (APAA), which directly multiplies a scaling factor to the gradient of loss function instead of using the sign function to normalize. The scaling factor can either be elaborately selected manually, or adaptively generated according to different image characteristics. Specifically, to further take the characteristics of different images into consideration, we propose an adaptive scaling factor generator to automatically generate the suitable scaling factor in each attack step during the generation of adversarial examples. 
  From \cref{fig:motivation}, we can clearly see that, since our method can adaptively adjust the step size in each attack step, a large step size can be used in the first few steps of iterative attacks to make full use of the perturbation budget, and when close to the global optimal point, the step size can be adaptively reduced. With more accurate update direction, our APAA may reach the global optimum with fewer update steps and perturbations, which means the generated adversarial examples are more aggressive and thus can improve the corresponding attack success rate.

  Adversarial examples have an intriguing property of transferability, where adversarial examples generated by one model can also fool other unknown models. Wang \etal~\cite{wang2021a} discover the negative correlation between the adversarial transferability and the interaction inside adversarial perturbations. Based on their findings, we theoretically demonstrate that the adversarial examples generated by our APAA method have better black-box transferability.

  The main contributions of our work are as follows:

  1. We propose an effective attack method by directly multiplying the gradient of loss function by a scaling factor instead of employing the sign function, which is nearly used in all existing gradient-based methods to normalize the gradient.

  2. We propose an adaptive scaling factor generator to generate the suitable scaling factor in each attack step according to the characteristics of different images, which gets rid of manually hyperparameter searching.

  3. We theoretically demonstrate that the adversarial examples generated by our method achieve better black-box transferability.

  4. Extensive experiments on the datasets of CIFAR10 and ImageNet show the superiority of our proposed method. Our method significantly improves the transferability of generated adversarial examples and achieves higher attack success rates on both white-box and black-box settings than the state-of-the-art methods with fewer update steps and perturbations budgets.

\section{Related Work}
\label{sec:related}
  The phenomenon of adversarial examples is proposed by Szegedy \etal~\cite{szegedy2014intriguing}. The attack and defense methods promote the development of each other in recent years, which are briefly reviewed in this section, respectively.

\subsection{Attack Methods}

  The attack methods mainly consist of the generative-based~\cite{xiao2018generating, zhao2018generating, song2018constructing, ameya2019semantic, qiu2020semantivadv, xiao2021improving} and gradient-based methods~\cite{goodfellow2015explaining, kurakin2017adversarial, madry2018towards, dong2018boosting, xie2019improving, dong2019evading, lin2020nesterov}, and the latter ones are the mainstream. 
  Our work mainly focuses on the gradient-based attack methods under the setting of $L_\infty$ norm. 
  Before introducing the methods in detail, we first introduce some notations that will be used later. Let $\bm{x}$ and $y$ be the original image and its corresponding class label, respectively. Let $J(\bm{x}, y)$ be the loss function of cross-entropy. Let $\bm{x_{adv}}$ be the generated adversarial example. Let $\epsilon$ and $\alpha$ be the total perturbation budget and the budget in each step of the iterative methods. $\Pi_{\bm{x}, \epsilon}$ means to clip the generated adversarial examples within the $\epsilon$-neighborhood of the original image on $L_\infty$ norm.

  \textbf{Fast Gradient Sign Method.} FGSM~\cite{goodfellow2015explaining} is a one-step method for white-box attack, which directly utilizes the gradient of loss function to generate the adversarial example:
  \begin{equation}
  \label{equ:fgsm}
  \bm{x^{adv}} = \bm{x} + \epsilon \cdot \mathrm{sign} (\nabla_{\bm{x}} J(f(\bm{x}), y)).
  \end{equation}
  
  \textbf{Basic Iterative Method.} BIM~\cite{kurakin2017adversarial} is an extension of FGSM, which uses the iterative method to improve the attack success rate of adversarial examples:
  \begin{equation}
    \bm{x^{adv}_{t+1}} = \Pi_{\bm{x}, \epsilon}(\bm{x^{adv}_t} + \alpha \cdot \mathrm{sign} (\nabla_{\bm{x}} J(f(\bm{x^{adv}_t}), y))),
  \end{equation}
  where $\bm{x^{adv}_0} = \bm{x}$ and the subscript $t$ is the index of iteration.
  
  \textbf{Momentum Iterative Fast Gradient Sign Method.} MIFGSM~\cite{dong2018boosting} proposes a momentum term to accumulate the gradient in previous steps to achieve more stable update directions, which greatly improves the transferability of generated adversarial examples:
  \begin{gather}
    \bm{g_{t+1}} = \mu \cdot \bm{g_t} + \frac{\nabla_{\bm{x}} J(f(\bm{x^{adv}_t}), y)}{\| \nabla_{\bm{x}} J(f(\bm{x^{adv}_t}), y)\|_1}, \label{eqn:g_t} \\
    \bm{x^{adv}_{t+1}} = \Pi_{\bm{x}, \epsilon}(\bm{x^{adv}_t} + \alpha \cdot \mathrm{sign}(\bm{g_{t+1}})),
  \end{gather}
  where $\bm{g_t}$ denotes the momentum item of gradient in the $t$-th iteration and $\mu$ is a decay factor.

  Since then, various methods have been further proposed to improve the transferability of adversarial examples. A randomization operation of random resizing and zero-padding to the original image is proposed in \textbf{DIM}~\cite{xie2019improving}.
  \textbf{TIM}~\cite{dong2019evading} proposes a translation-invariant attack method by convolving the gradient with a Gaussian kernel to further improve the transferability of adversarial examples.
  Inspired by Nesterov accelerated gradient~\cite{Nesterov1983AMF}, \textbf{SIM}~\cite{lin2020nesterov} amends the accumulation of the gradients to effectively look ahead and improve the transferability of adversarial examples. In addition, SIM also proposes to use several copies of the original image with different scales to generate the adversarial example.
  \textbf{SGM}~\cite{wu2020skip} finds that using more gradients from the skip connections rather than the residual modules can craft adversarial examples with higher transferability.
  \textbf{VT}~\cite{wang2021enhancing} considers the gradient variance of the previous iteration to tune the current gradient so as to stabilize the update direction and escape from poor local optima.
  \textbf{EMI}~\cite{wang2021boosting} accumulates the gradients of data points sampled in the gradient direction of previous iteration to find more stable direction of the gradient.
  \textbf{IR}~\cite{wang2021a} discovers the negative correlation between the adversarial transferability and the interaction inside adversarial perturbations and proposes to directly penalize interactions during the attacking process, which significantly improves the adversarial transferability.
  \textbf{AIFGTM}~\cite{zou2022making} also considers the limitations of the basic sign structure and proposes an ADAM iterative fast gradient tanh method to generate indistinguishable adversarial examples with high
  transferability.
    
\subsection{Defense Methods}
  Adversarial defense aims to improve the robustness of the target model in the case of adversarial examples being the inputs. The defense methods can mainly be categorized into adversarial training, input transformation, model ensemble, and certified defenses. The adversarial training methods~\cite{madry2018towards, tramer2017ensemble, song2020robust, pang2020boosting, wong2020fast} use the adversarial examples as the extra training datas to improve the robustness of the model. The input transformation methods~\cite{dziugaite2016study, liao2018defense, samangouei2018defense, liu2019feature, jia2019comdefend} tend to denoise the adversarial examples before feeding them into the classifier. The model ensemble methods~\cite{liu2018towards, pang2019improving, yang2020dverge} use multiple models simultaneously to reduce the influence of adversarial examples on the single model and achieve more robust results. Certified defense methods~\cite{raghunathan2018certified, wong2018scaling, cohen2019certified, jia2020certified} guarantee that the target model can correctly classify the adversarial examples within the given distance from the original images.

\section{Method}
\label{sec:method}
  In this section, we first analyze the defect of the sign function used in existing gradient-based attack methods in \cref{sec:rethinking}. Then we propose our method of Adaptive Perturbation for Adversarial Attack (APAA) in \cref{sec:APAA}, \ie, utilizing a scaling factor to multiply the gradient instead of the sign function normalization. The scaling factor can either be elaborately selected manually (in \cref{sec:fixed}), or adaptively generated according to different image characteristics by a generator (in \cref{sec:adaptive}). Finally, we theoretically demonstrate that our proposed method can improve the black-box transferability of adversarial examples in \cref{sec:theory}.

\subsection{Rethinking the Sign Function}
\label{sec:rethinking}
  The task of adversarial attack is to do a minor modification on the original images with human-imperceptible noises to fool the target model, \ie, misclassifying the adversarial examples. The gradient-based methods generate the adversarial examples by maximizing the cross-entropy loss function, which can be formulated as follows:
  \begin{equation}
    \mathop{\arg\max}_{\bm{x^{adv}}} J(f(\bm{x^{adv}}), y), \quad \mathrm{s.t.} \|\bm{x^{adv}} - \bm{x}\|_{p} \leq \epsilon,
  \end{equation}
  where $p$ could be $0,1,2$ and $\infty$.
  
  Since the gradient varies in different pixels, the sign function is usually used to normalize the gradient, which is convenient to set the step size of each step during the attack (\eg, \cref{equ:fgsm}). Under the most commonly used $L_\infty$ norm setting in the adversarial attack, \ie, constraining the maximal $L_\infty$ norm of the generated adversarial perturbations, the use of the sign function can also leverage the largest perturbation budget to enhance the aggressiveness of adversarial examples.

  The sign function normalizes all values of the gradient into $\{0, +1, -1\}$. Although this method can scale the gradient value to make full use of the perturbation budget in the $L_\infty$ attack, the resultant update directions in adversarial attack are limited, \eg, there are only eight possible update directions in the case of a two-dimensional space ((0, 1), (0, -1), (1, 1), (1, -1), (1, 0), (-1, -1), (-1, 0), (-1, 1)).

  We use a two-dimensional toy example (as shown in~\cref{fig:motivation}) to demonstrate the limitation of the existing attack methods with the sign function. The loss function in~\cref{fig:motivation} is a Gaussian mixture model of the following expression:
  \begin{align}
    f(x,y) = &\exp\{-[(x+2.8)^2+2(y-2)^2+0.5(x+2.8)(y-2)]\}  \notag    \\
     +0.7 &\exp\{-[(x-1)^2+(y-1)^2-0.8(x-1)(y-1)]\} \notag \\
     - &\exp\{-[(x+3)^2+(y+2)^2-0.5(x+3)(y+2)]\}. \label{eqn:example}
  \end{align}
  From the figure, we can clearly see that the attack directions in sign-based methods (\eg, BIM~\cite{kurakin2017adversarial}) are distracted and not accurate anymore, which conversely needs more update steps and perturbations budgets to implement a successful attack. The inaccurate update direction may cause the generated adversarial examples to be sub-optimal.

\subsection{Adaptive Perturbation for Adversarial Attack}
\label{sec:APAA}
  To solve the problem mentioned above, we propose a method of Adaptive Perturbation for Adversarial Attack (APAA). Specifically, we propose to directly multiply the gradient by a scaling factor instead of normalizing it with the sign function. The scaling factor can be determined either elaborately selected manually, or adaptively achieved from a generator according to different image characteristics. We will introduce each of them in the following, respectively.

\subsubsection{Fixed Scaling Factor}
\label{sec:fixed}
  First, we propose to directly multiply the gradient by a fixed scaling factor, which not only maintains the accurate gradient directions but also can flexibly utilize the perturbation budget through the adjustment of the scaling factor:
  \begin{equation}
    \label{equ:gamma}
    \bm{x^{adv}_{t+1}} = \Pi_{\bm{x}, \epsilon}(\bm{x^{adv}_t} + \gamma \cdot \nabla_{\bm{x}} J(f(\bm{x^{adv}_t}), y)),
  \end{equation}
  where $\gamma$ is the scaling factor, $\Pi_{\bm{x}, \epsilon}$ means to clip the generated adversarial examples within the $\epsilon$-neighborhood of the original image on $L_\infty$ norm.

  Our method is easy to implement and can be combined with all existing gradient-based attack methods (\eg, MIFGSM~\cite{dong2018boosting}, DIM~\cite{xie2019improving}, TIM~\cite{dong2019evading}, SIM~\cite{lin2020nesterov}). We take the MIFGSM method as an example. When integrating with MIFGSM, the full update formulation is as follows:
  \begin{gather}
    \bm{x_0^{adv}} = \bm{x}, \\
    \bm{g_{t+1}} = \mu \cdot \bm{g_t} + \frac{\nabla_{\bm{x}} J(f(\bm{x^{adv}_t}), y)}{\| \nabla_{\bm{x}} J(f(\bm{x^{adv}_t}), y)\|_1}, \label{eqn:g} \\
    \bm{x^{adv}_{t+1}} = \Pi_{\bm{x}, \epsilon} (\bm{x_t^{adv}} + \gamma \cdot \bm{g_{t+1}}),
  \end{gather}
  where $\mu$ is the decay factor in MIFGSM, $\gamma$ is the scaling factor.

  \begin{figure*}[t]
    \centering
    \includegraphics[width=0.9\textwidth]{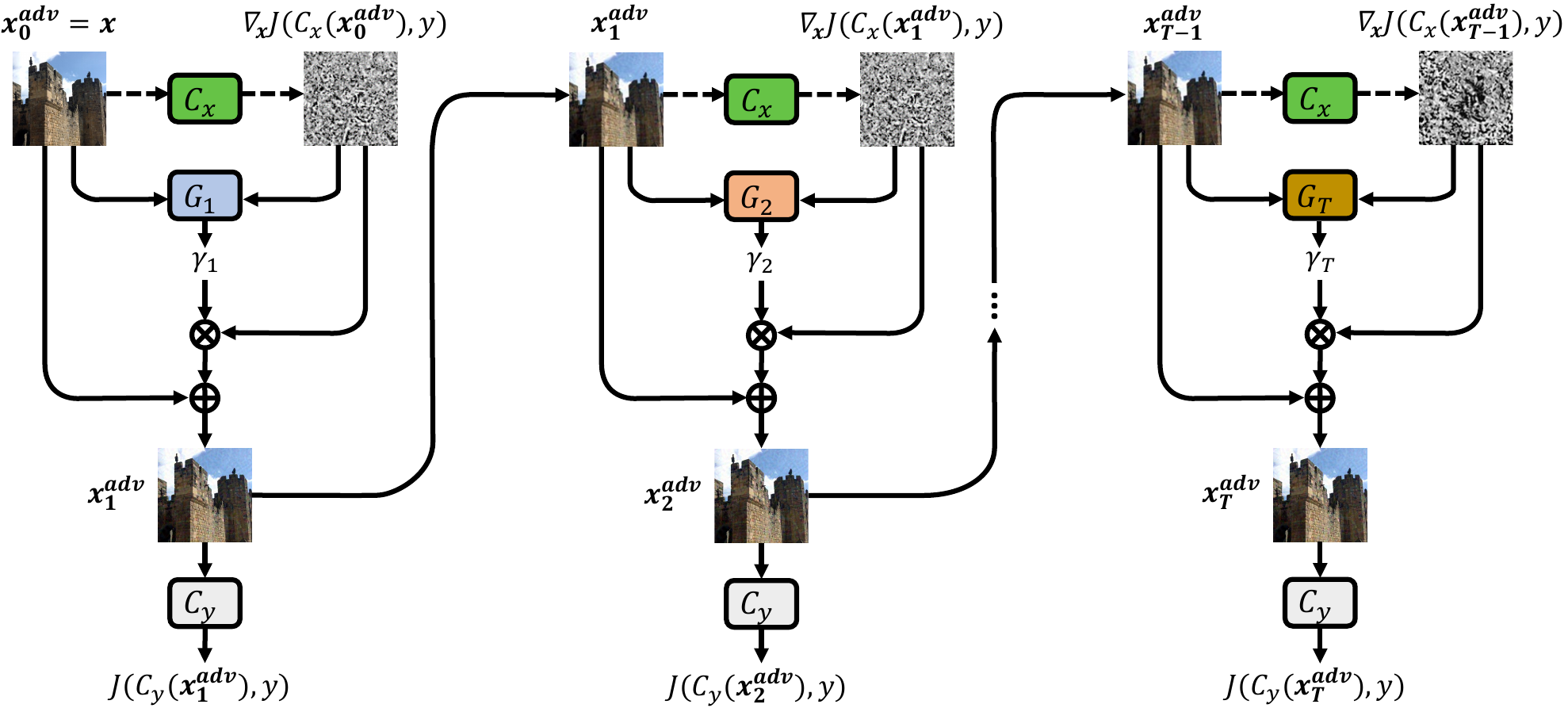}
    \caption{The overview of the scaling factor generator. In each step $t$, the gradient of one randomly selected model $C_x$ is first calculated. Simultaneously, the adversarial example from the previous step together with the gradient information are fed into the generator $G_t$ to generate the scaling factor $\gamma_t$. Then the scaling factor and the gradient information are used to generate the adversarial perturbation. The adversarial examples generated in all steps are all fed into another classifier $C_y$ to optimize the parameters of generator $G_t$ through maximizing the cross-entropy loss between the output classification probability and the ground truth label.}
    \label{fig:model}
  \end{figure*}

\subsubsection{Adaptive Scaling Factor}
\label{sec:adaptive}
  Considering that suitable scaling factors vary across different images, we further propose a generator to directly generate the adaptive scaling factor in each step of adversarial examples generation, which also gets rid of manually hyperparameter searching.
  
  The overview of our proposed adaptive scaling factor generator is shown in \cref{fig:model}. We employ an iterative method to generate the adversarial example with $T$ steps.
  In each step $t$, the gradient of the adversarial examples generated in the previous step is calculated through a randomly selected white-box model $C_x$. The corresponding scaling factor $\gamma_t$ is also generated through the generator $G_t$ by feeding the adversarial example together with the gradient information as the input. Then the gradient information and scaling factor are used to generate the new adversarial perturbation. We used the BIM-style update method ($\bm{x_t^{adv}} = \bm{x_{t-1}^{adv}} + \gamma_t \cdot \nabla_{\bm{x}} J(C_x(\bm{x_{t-1}^{adv}}), y)$) as an example in the figure for convenience, but in fact any existing gradient-based attack method can be utilized. The generated adversarial examples in all steps are used to calculate the cross-entropy loss by another classifier $C_y$. Then the parameters in the generator $G_t$ are updated by maximizing the cross-entropy loss through backpropagation, aiming to generate the adversarial examples, which can mislead the classifier. The generators used in each step share the same architecture but do not share the parameters. The procedure of training the generator is summarized in \cref{alg:gen}.

  \begin{algorithm}
    \algnewcommand\algorithmicinput{\textbf{Input:}}
    \algnewcommand\Input{\item[\algorithmicinput]}
    \algnewcommand\algorithmicoutput{\textbf{Output:}}
    \algnewcommand\Output{\item[\algorithmicoutput]}
    \caption{The training of adaptive scaling factor generator}
    \label{alg:gen}
    \begin{algorithmic}[1]
        \Input the total training step $N$
        \Input the number of attack iteration $T$
        \Input the learning rate $\beta$
        \Input the dataset $(\mathcal{X, Y})$
        \Input the white-box models $C_1, \cdots, C_n$
        \Output the scaling factor generators $G_1, G_2, \cdots, G_T$ with the parameters of $\bm{\theta}$
        
        \For {i $\in \{1, \cdots N\}$ }
          \State Sample the image and label pair $(\bm{x}, y)$ from $(\mathcal{X, Y})$
          \State Randomly choose two classifiers $C_x$ and $C_y$ from $C_1, \cdots, C_n$
          \State $\bm{x_0^{adv}} = \bm{x}$
          \For {t $\in \{1, \cdots T\}$ }
            \State $\bm{grad_t}=\nabla_{\bm{x}} J(C_x(\bm{x_{t-1}^{adv}}), y)$
            \State $\gamma_t=G_t(\bm{x_{t-1}^{adv}}, \bm{grad_t})$
            \State $\bm{x_t^{adv}} = \Pi_{\bm{x}, \epsilon} (\gamma_t \cdot \bm{grad_t} + \bm{x_{t-1}^{adv}})$ \Comment{using the update in BIM as an example} 
            \State $loss_t=J(C_y(\bm{x_{t}^{adv}}), y)$
            \State $G_t(\bm{\theta}) = G_t(\bm{\theta}) + \beta \cdot \nabla_{\bm{\theta}} (loss_t)$
          \EndFor
        \EndFor
    \end{algorithmic}
  \end{algorithm}

  From \cref{fig:motivation}, we can observe the difference between our proposed APAA method and existing sign-based methods (\eg, BIM~\cite{kurakin2017adversarial}) in generating adversarial examples. To show the generating process, we set the initial point at (-2.900, -1.850) for both our APAA and BIM, and the step size for BIM to be 0.5. For our proposed APAA, a single-layer neural network is trained to dynamically generate the adaptive scaling factor, considering the simplicity of this toy example. In the figure, the orange and blue lines depict the process of generating adversarial examples using BIM and APAA, respectively. As can be seen from the figure, BIM finally reaches the point (1.003, 0.752) after 12 steps, approximating a local optimum (1.000,1.000), while only 9 steps are needed for APAA to reach the point (-2.730, 2.033), very close to the global optimum (-2.800, 2.000). The reason behind is that our APAA enables a more accurate update direction and allows adaptive step size, while the update direction in BIM is con-strained and lacks precision. In our APAA, a large step size can be used in the first few steps of iterative attacks to make full use of the perturbation budget, and when close to the global optimal point, the step size can be adaptively reduced. Without loss of generality, we believe our APAA has a higher probability of reaching the global optimum in fewer steps under real scenarios.

  It is worth noting that two methods proposed above (\ie, APAA with fixed scaling factor and APAA with adaptive scaling factor) are complementary, each with its own distinct advantages. So the attacker can consider to choose which one to use depending on the specific situation. Specifically, as shown in the experiment part, APAA with adaptive scaling factor performs better and can automatically obtain adaptive scaling factors for each step in the attack but requires training a scaling factor generator. However, due to the lightweight structure of the generator (as illustrated in~\cref{tab:arch}), the extra inference time brought by the generator during the generation of adversarial examples is negligible. On the other hand, APAA with fixed scaling factor is simple, does not require training a generator network, and outperforms existing methods significantly, making it an excellent choice as well. Therefore, attackers can select one of them based on the specific scenario when conducting the attack.

\subsection{Theoretical Analysis}
\label{sec:theory}
  Adversarial examples have an intriguing property of transferability, where adversarial examples generated by one model can also fool other unknown models. 
  In addition to the intuitive idea to illustrate that our method can provide more accurate attack directions, we also provide a theoretical proof to show that our proposed method can meanwhile improve the black-box transferability of adversarial examples. Wang \etal~\cite{wang2021a} utilizes the Shapley interaction index proposed in game theory~\cite{grabisch1999an, shapley1953} to analyze the interactions inside adversarial perturbations. Through extensive experiments, they discover the negative correlation between the adversarial transferability and the interaction inside adversarial perturbations. Based on their findings, we theoretically demonstrate that the adversarial examples generated by our APAA method have better black-box transferability.

  Shapley value~\cite{shapley1953} was first proposed in game theory in 1953. In a multiplayer game, players work together to obtain a high reward. Shapley Value is used to distribute the rewards shared by everyone according to each player's contribution fairly. $\Omega=\{1,2,\cdots,n\}$ represents the set of all players, $v(\cdot)$ represents the reward function, $\phi(i|\Omega)$ represents an unbiased estimate of the contribution of the i-th player to all players $\Omega$, which can be calculated as follows:
  \begin{equation}
    \phi(i|\Omega)=\sum_{S\subseteq \Omega \backslash \{i\}} \frac{|S|!(n-|S|-1)!}{n!}(v(S\cup\{i\})-v(S)).
  \end{equation}
  When applying the Shapley value into the adversarial examples, $\phi(i|\Omega)$ is used to measure the contribution of each perturbed pixel $i\in\Omega$ to the attack. During the adversarial examples, the reward function can be definded as:
  \begin{equation}
    v(S)=\max_{y'\neq y}h_{y'}(\bm{x}+\bm{\delta}^{(S)})-h_y(\bm{x}+\bm{\delta}^{(S)}),
  \end{equation}
  where $h_y(\cdot)$ represents the y-th element in the logits layer of DNN model, $y$ is the gound truth label of input image $\bm{x}$, $\bm{\delta}^{(S)}$ is the perturbation which only contains perturbation units in $S$, \ie, $\forall i \in S, \bm{\delta}_{i}^{(S)}=\bm{\delta}_i; \forall i \notin S, \bm{\delta}_i^{(S)}=0$. According to~\cite{grabisch1999an}, the Shapley interaction index between units $i$ and $j$ is defined as follows: 
  \begin{equation}
    I_{ij}(\delta)=\phi(S_{ij}|\Omega')-[\phi(i|\Omega\backslash \{j\})+\phi(j|\Omega\backslash\{i\})],
  \end{equation}
  where $S_{ij}=\{i,j\}$ regards perturbation units $i,j$ as a singleton unit, $\Omega'=\Omega\backslash\{i,j\}\cap S_{ij}$, $\phi(S_{ij}|\Omega')$ is the joint contribution of $S_{ij}$.

  Wang \etal~\cite{wang2021a} utilizes $\mathbb{E}_{a,b}[I_{ab}(\bm{\delta})]$ to estimate the interaction inside perturbations. Through extensive experiments, Wang \etal~discover the negative correlation between the transferability and interactions, \ie, the adversarial examples with smaller interactions have the better black-box transferability. In the following, we take MIFGSM~\cite{dong2018boosting} as an example to prove that the adversarial examples generated by our APAA method have smaller interaction values, which also confirms that our method has better black-box transferability.

  To simplify the proof, we do not consider some tricks in the adversarial attack, such as gradient normalization and the clip operation. The MIFGSM method combined with APAA can be formulated as:
  \begin{gather}
    \bm{g_{t}}= \mu \cdot \bm{g_{t-1}}+g(\bm{x}+\bm{\delta_{t-1}}), \label{equ:gt}\\
    \bm{\delta_{t}} = \sum_{i=1}^t \gamma \cdot \bm{g_{t}}, \label{equ:deltat}
  \end{gather}
  where $g(\bm{x})=\frac{\partial L(\bm{x})}{\partial \bm{x}}$.
  \newtheorem{prop}{Proposition}
    \begin{prop}[The perturbations generated by MIFGSM with APAA]
      \label{prop:1}
      The adversarial perturbation generated by MIFGSM with APAA at $m$-th step is given as:
      \begin{align}
        \bm{g_{m}}=& a_m \cdot \bm{g} + b_m \cdot \gamma \bm{Hg}, \label{equ:gm}\\
        \bm{\delta_{m}}=& c_m \cdot \gamma \bm{g} + d_m \cdot \gamma^2 \bm{Hg}, \label{equ:deltam}
      \end{align}
      where $\bm{g}$ and $\bm{H}$ are the first and second order gradients of $L(\bm{x})$ with respect to $\bm{x}$, respectively,
      \begin{align}
        a_m=&\sum_{i=1}^m \mu^{i-1}, \label{equ:am} \\ 
        b_m=&\sum_{i=1}^m (m-i+1)(i-1) \mu^{i-2}, \label{equ:bm}\\
        c_m=&\sum_{i=1}^m (m-i+1) \mu^{i-1}, \label{equ:cm}\\
        d_m=&\sum_{i=1}^m \frac{(m-i+2)(m-i+1)(i-1)}{2} \mu^{i-2}. \label{equ:dm}
      \end{align}
    \end{prop}
  Detailed proofs are provided in the appendix (\cref{sec:prop1}).

  Further, we can calculate the interaction inside perturbations generated by MIFGSM with APAA.
    \begin{prop}[The interaction inside perturbations generated by MIFGSM with APAA]
      \label{prop:2}
      The interaction inside adversarial perturbations generated by MIFGSM with APAA at $m$-th step is given as:
      \begin{equation}
        \mathbb{E}_{a,b}(I_{ab};\gamma) = A\gamma^2+2B\gamma^3, \label{equ:interaction}
      \end{equation}
      where
      \begin{equation*}
        A=\mathbb{E}_{a,b}[c_m^2 \bm{g}_a\bm{g}_b\bm{H}_{ab}], \quad B=\mathbb{E}_{a,b}[c_md_m\bm{g}_a\bm{H}_{ab}\bm{g}^\top\bm{H}_{*b}]\geq 0,
      \end{equation*}
      $\bm{g}$ and $\bm{H}$ are the first and second order gradients of $L(\bm{x})$ with respect to $\bm{x}$, respectively, $\bm{g}_a$ and $\bm{g}_b$ are the $a$-th and $b$-th elements in $\bm{g}$, $\bm{H}_{*b}$ represents the $b$-th column of Hessian matrix $\bm{H}$.
    \end{prop}
  Detailed proofs are provided in the appendix (\cref{sec:prop2}).

  Through experiments, we find that the magnitude of gradients obtained by derivation in the MIFGSM method is predominately smaller than $10^{-2}$. Given that the gradient values are usually small, normalizing the gradient with the sign function can be approximated as multiplying the gradient by a rather large coefficient. When we convert these gradients to $\{-1, 1\}$ with the sign function, it is equivalent to multiplying by a $\gamma_{MIFGSM}$ on the order of greater than $10^2$. On the other hand, as evident from the experiments that follows, our $\gamma_{APAA}$ takes values below 10 (0.4 or 0.8 on ImageNet and 8 on CIFAR10). Hence, we can conclude that $0<\gamma_{APAA} \ll \gamma_{MIFGSM}$. We treat \cref{equ:interaction} as a cubic function of $\gamma$, taking into account that coefficient $B$ is greater than 0, so we can achieve $\mathbb{E}_{a,b}(I_{ab};\gamma_{APAA}) < \mathbb{E}_{a,b}(I_{ab};\gamma_{MIFGSM})$, which means our APAA method have better black-box transferability.
  \begin{figure}[t]
    \centering
    \includegraphics[width=0.9\columnwidth]{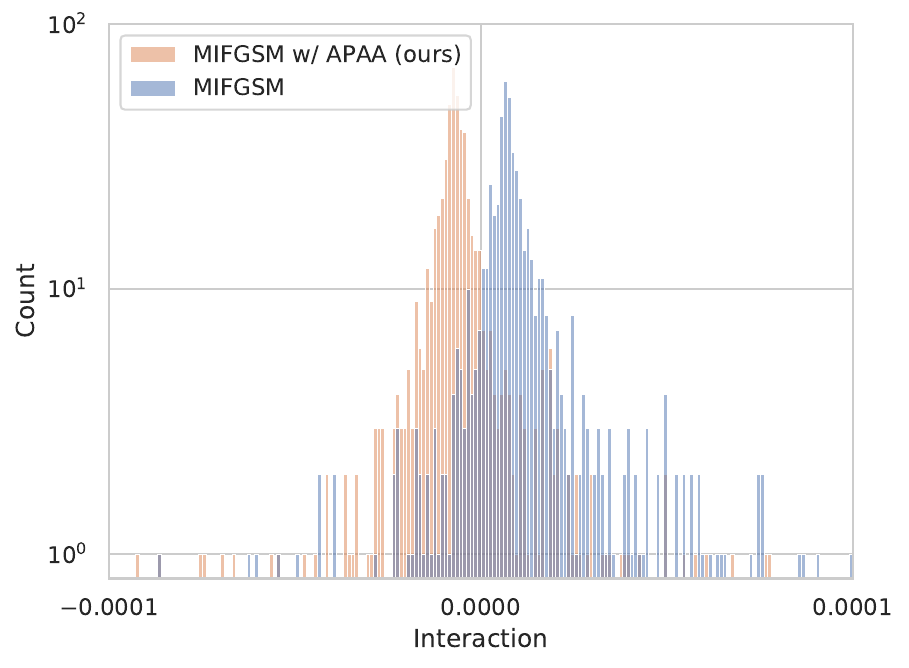}
    \caption{The comparison of histograms of the interaction inside perturbations generated by our APAA and that of MIFGSM~\cite{dong2018boosting}. Smaller interaction values correspond to better black-box transferability. The experiment is conducted on 1000 images of CIFAR10 testset.}
    \label{fig:interaction}
  \end{figure}
  Moreover, we also verify our proof by experiments. As shown in \cref{fig:interaction}, we can clearly see that the interaction inside perturbations generated by our APAA is significantly smaller than that of MIFGSM.
  Combined with the negative correlation between the adversarial transferability and the interaction inside adversarial perturbations, our proposed APAA method is theoretically proven to have better black-box transferability.

\section{Experiments}
  We first introduce the setting of experiments in \cref{sec:setting}. Then we demonstrate the effectiveness of our proposed APAA with fixed scaling factor in \cref{sec:exp_fixed}. Finally, we conduct several experiments to show the superiority of the adaptive scaling factor generator in \cref{sec:exp_adaptive}. We denote the APAA with the fixed scaling factor as $\text{APAA}_f$ and the APAA with the adaptive scaling factor generator as $\text{APAA}_a$.
  \begin{table}[!th]
    \begin{center}
        \caption{The hyperparameters used in baseline methods on ImageNet and CIFAR10 datasets. For the dataset of ImageNet, we use the hyperparameters reported in origin papers. For the dataset of CIFAR10, we find that the setting used in ImageNet is ineffective and determine the optimal value by hyperparameter searching.}
        \label{tab:baseline}
        \resizebox{\columnwidth}{!}{
          \begin{tabular}{c|c|c}
              \hline
              ~ & ImageNet & CIFAR10 \\
              \hline
              MIFGSM & $\mu=1.0$ &  $\mu=1.0$ \\
              DIM & $p=0.7$ & $p=0.7$\\
              TIM &  $k=15$ & $k=3$ \\
              SIM & $m=5$ & $m=2$ \\
              VT & $N=20$, $\beta=1.5$ & \\
              EMI & $N=11$, $\eta=7$ & \\
              SGM & $\gamma=0.2$ for Res-152, $\gamma=0.5$ for Dense-201 & \\
              IR &$\lambda=1$ for Res-34, $\lambda=2$ for Dense-121 & \\
              AIFGTM & $\lambda=1.3$, $\mu_1=1.5$, $\mu_2=1.9$, $\beta_1=0.9$, $\beta_2=0.99$ & \\
              SVRE & & $M=8$ \\
              \hline
          \end{tabular}
        }
    \end{center}
  \end{table}

  \begin{table}[!th]
    \begin{center}
        \caption{The architecture of the adaptive scaling factor generator. The $k$, $s$ and $p$ in the Conv2d layer means the kernel size, stride and padding, respectively.}
        \label{tab:arch}
        \resizebox{\columnwidth}{!}{
          \begin{tabular}{c|c|c|c}
              \hline
              No. & Layer & Input Shape & Output Shape \\
              \hline
              1 & Conv2d(k=3, s=2, p=1) & [32,32,6] & [16,16,32] \\
              2 & InstanceNorm2d & [16,16,32] & [16,16,32] \\
              3 & Conv2d(k=3, s=2, p=1) & [16,16,32] & [8,8,32] \\
              4 & InstanceNorm2d & [8,8,32] & [8,8,32] \\
              5 & Conv2d(k=3, s=2, p=1) & [8,8,32] & [4,4,32] \\
              6 & InstanceNorm2d & [4,4,32] & [4,4,32] \\
              7 & Flatten & [4,4,32] & [512] \\
              8 & Linear & [512] & [128] \\
              9 & Linear & [128] & [1] \\
              \hline
          \end{tabular}
        }
    \end{center}
  \end{table}

\subsection{Experiment Setting}
\label{sec:setting}
  \textbf{Datasets.} We use ImageNet~\cite{russakovsky2015imagenet} and CIFAR10~\cite{krizhevsky2009learning} datasets to conduct the experiments. CIFAR10 has 50000 training images and 10000 test images in different classes. For ImageNet, we use two sets of subsets\footnote{\url{https://github.com/cleverhans-lab/cleverhans/tree/master/cleverhans_v3.1.0/examples/nips17_adversarial_competition/dataset}\label{foot:subset1}}\textsuperscript{,}\footnote{\url{https://drive.google.com/drive/folders/\\1CfobY6i8BfqfWPHL31FKFDipNjqWwAhS}\label{foot:subset2}} in the ImageNet dataset~\cite{russakovsky2015imagenet} to conduct experiments. Each set contains 1000 images, covering almost all categories in ImageNet, which has been widely used in previous works. The maximation perturbation budgets on CIFAR10 and ImageNet are 8 and 16 with $L_\infty$ norm under the scale of 0-255, respectively.

  \begin{table*}[!bh]
    \begin{center}
      \centering
      \caption{The ablation study of the hyperparameter $\gamma$ in~\cref{equ:gamma}. The result is the attack success rates of the adversarial examples on \textbf{ImageNet} under \textbf{untargeted attack} setting. The source model used in the experiment is IncRes-v2. The bold coefficients indicate the values we finally selected in the respective method.}
      \label{tab:imagenet_gamma}
      \begin{subtable}[t]{\textwidth}
        \caption{The evaluation on the normally trained models.}
        \centering
        \resizebox{0.9\textwidth}{!}{
          \begin{tabular}{l|ccccccccc|cc}
          \hline
          \multicolumn{1}{c|}{\multirow{2}{*}{Method}} & \multicolumn{9}{c}{Target Model} & \multicolumn{2}{|c}{Distance Metric} \\ \cline{2-12}
          \multicolumn{1}{c|}{}                        & IncRes-v2     & Inc-v3         & Inc-v4        & Res-101       & Res-152 & Mob\textsubscript{1.0}    & Mob\textsubscript{1.4}    & PNAS    & NAS  & MAD              & RMSD      \\ \hline
          TIM~\cite{dong2019evading}              & 98.3  & 86.5  & 83.9  & 73.3  & 73.8  & 81.4  & 83.7  & 75.1  & 77.3 &  10.302 & 11.107    \\
          TIM w/ APAA$_f$ ($\gamma$=0.2)          & 97.7  & 84.6  & 82.2  & 71.3  & 72.6  & 76.4  & 80.0  & 72.4  & 73.7 &  5.993 & 6.668    \\                            
          \textbf{TIM w/ APAA$_f$ ($\gamma$=0.4)}          & 98.6  & 88.6  & 85.4  & 75.3  & 76.1  & 83.7  & 87.9  & 77.2  & 80.2 &  8.887 & 9.693    \\                            
          TIM w/ APAA$_f$ ($\gamma$=0.6)          & 99.0  & 88.8  & 85.7  & 75.7  & 75.9  & 85.7  & 88.2  & 76.0  & 80.3 &  10.497 & 11.288    \\                       
          TIM w/ APAA$_f$ ($\gamma$=0.8)          & 99.3  & 90.0  & 86.7  & 75.0  & 75.4  & 87.5  & 88.6  & 77.5  & 80.2 &  11.453 & 12.195    \\                        
          TIM w/ APAA$_f$ ($\gamma$=1.0)          & 99.4  & 89.7  & 86.9  & 75.3  & 75.8  & 85.9  & 88.9  & 79.2  & 80.1 &  12.092 & 12.785    \\   \hline
          EMI~\cite{wang2021boosting}             & 99.5  & 95.1  & 93.0  & 88.0  & 87.3  & 90.8  & 93.8  & 89.8  & 92.0 &  10.479 & 11.282    \\
          EMI w/ APAA$_f$ ($\gamma$=0.2)          & 99.6  & 96.0  & 94.3  & 88.7  & 88.4  & 91.4  & 93.8  & 89.7  & 91.2 &  6.335 & 7.074    \\      
          \textbf{EMI w/ APAA$_f$ ($\gamma$=0.4)}          & 99.6  & 97.0  & 95.9  & 90.2  & 90.5  & 93.6  & 95.2  & 91.8  & 92.7 &  9.390 & 10.234    \\    
          EMI w/ APAA$_f$ ($\gamma$=0.6)          & 99.7  & 97.0  & 95.7  & 90.3  & 90.0  & 94.3  & 96.0  & 91.1  & 93.5 &  10.940 & 11.737    \\  
          EMI w/ APAA$_f$ ($\gamma$=0.8)          & 99.9  & 96.9  & 95.8  & 89.5  & 89.6  & 93.8  & 97.3  & 91.8  & 94.3 &  11.845 & 12.576    \\
          EMI w/ APAA$_f$ ($\gamma$=1.0)          & 99.8  & 97.1  & 95.0  & 89.5  & 89.1  & 93.4  & 96.6  & 92.0  & 93.0 &  12.416 & 13.088    \\   \hline
          VT~\cite{wang2021enhancing}             & 98.5  & 89.1  & 87.0  & 79.8  & 80.4  & 82.6  & 86.6  & 80.5  & 82.2 &  10.031 & 10.810     \\
          VT w/ APAA$_f$ ($\gamma$=0.2)           & 97.2  & 82.7  & 81.4  & 71.2  & 72.3  & 74.1  & 77.5  & 71.8  & 75.2 &  4.193 & 4.703    \\                            
          VT w/ APAA$_f$ ($\gamma$=0.4)           & 98.9  & 89.5  & 88.3  & 80.2  & 80.1  & 82.7  & 85.6  & 81.9  & 82.2 &  6.742 & 7.452    \\                            
          VT w/ APAA$_f$ ($\gamma$=0.6)           & 99.1  & 91.9  & 89.7  & 80.9  & 82.9  & 85.4  & 88.6  & 82.4  & 84.6 &  8.369 & 9.144    \\                            
          \textbf{VT w/ APAA$_f$ ($\gamma$=0.8)}           & 99.5  & 92.3  & 90.9  & 82.6  & 82.9  & 87.5  & 90.6  & 83.8  & 85.4 &  9.513 & 10.298    \\                            
          VT w/ APAA$_f$ ($\gamma$=1.0)           & 99.4  & 93.2  & 90.4  & 83.3  & 82.8  & 87.8  & 90.4  & 83.4  & 85.6 &  10.346 & 11.117    \\   \hline
          AIFGTM~\cite{zou2022making}             & 97.4  & 82.2  & 78.0  & 72.5  & 72.4  & 77.1  & 80.4  & 74.0  & 75.1 &  10.415 & 11.269    \\
          AIFGTM w/ APAA$_f$ ($\gamma$=0.2)       & 90.4  & 59.2  & 54.1  & 49.5  & 49.7  & 53.4  & 56.5  & 50.7  & 52.5 &  3.823 & 4.209    \\                          
          AIFGTM w/ APAA$_f$ ($\gamma$=0.4)       & 96.2  & 78.2  & 75.7  & 67.5  & 67.2  & 72.6  & 77.0  & 70.4  & 72.8 &  6.968 & 7.699    \\                           
          AIFGTM w/ APAA$_f$ ($\gamma$=0.6)       & 98.1  & 84.1  & 80.9  & 74.0  & 73.5  & 79.7  & 82.2  & 76.0  & 78.0 &  8.121 & 8.982    \\                          
          \textbf{AIFGTM w/ APAA$_f$ ($\gamma$=0.8)}       & 98.3  & 84.2  & 81.8  & 74.2  & 74.0  & 80.5  & 84.5  & 78.2  & 79.7 &  9.004 & 9.742    \\                         
          AIFGTM w/ APAA$_f$ ($\gamma$=1.0)       & 97.9  & 86.3  & 83.1  & 75.9  & 76.4  & 82.0  & 84.9  & 79.2  & 80.4 &  11.268 & 12.075    \\   \hline
          \end{tabular}
        }
      \end{subtable}
      \begin{subtable}[t]{\textwidth}
        \caption{The evaluation on the defense models.}
        \resizebox{\textwidth}{!}{
          \begin{tabular}{l|cccccccccccc}
          \hline
          \multicolumn{1}{c|}{\multirow{2}{*}{Method}} & \multicolumn{12}{c}{Target Model}                                                 \\ \cline{2-13} 
          \multicolumn{1}{c|}{}                        & Inc-v3\textsubscript{adv} & Inc-v3\textsubscript{ens3} & Inc-v3\textsubscript{ens4} & IncRes-v2\textsubscript{ens} & HGD  & R\&P & NIPS-r3 & Bit-Red & JPEG & FD   & ComDefend & RS   \\ \hline
          TIM~\cite{dong2019evading}            & 56.4  & 64.7  & 55.7  & 51.6  & 60.6  & 53.0  & 59.1  & 42.4  & 75.3  & 67.2  & 70.3  & 36.2         \\
          TIM w/ APAA$_f$ ($\gamma$=0.2)        & 69.5  & 68.7  & 62.1  & 62.6  & 65.2  & 61.8  & 62.0  & 40.7  & 73.9  & 66.2  & 66.0  & 34.2         \\                                    
          \textbf{TIM w/ APAA$_f$ ($\gamma$=0.4)}        & 68.2  & 66.2  & 60.3  & 54.0  & 62.2  & 54.9  & 61.0  & 43.9  & 76.6  & 69.7  & 71.3  & 40.5         \\                                    
          TIM w/ APAA$_f$ ($\gamma$=0.6)        & 60.7  & 60.7  & 54.3  & 46.4  & 54.4  & 48.9  & 56.4  & 42.1  & 76.4  & 71.0  & 71.8  & 40.4         \\                                    
          TIM w/ APAA$_f$ ($\gamma$=0.8)        & 58.4  & 58.0  & 49.6  & 40.3  & 46.3  & 42.5  & 51.3  & 41.4  & 74.7  & 69.6  & 70.3  & 41.4         \\                                    
          TIM w/ APAA$_f$ ($\gamma$=1.0)        & 57.1  & 55.5  & 47.9  & 37.7  & 41.5  & 39.6  & 47.9  & 39.5  & 74.8  & 68.2  & 70.7  & 42.5         \\   \hline
          EMI~\cite{wang2021boosting}           & 76.8  & 79.3  & 71.3  & 66.2  & 76.9  & 69.1  & 76.0  & 58.9  & 87.8  & 83.3  & 84.9  & 50.2         \\
          EMI w/ APAA$_f$ ($\gamma$=0.2)        & 88.1  & 85.8  & 81.0  & 79.9  & 84.8  & 81.0  & 82.9  & 59.4  & 88.5  & 83.9  & 84.6  & 47.3         \\ 
          \textbf{EMI w/ APAA$_f$ ($\gamma$=0.4)}        & 86.5  & 83.7  & 76.3  & 74.6  & 80.2  & 75.2  & 79.1  & 60.9  & 90.3  & 85.3  & 87.5  & 56.9         \\  
          EMI w/ APAA$_f$ ($\gamma$=0.6)        & 82.7  & 79.0  & 71.9  & 62.6  & 71.1  & 68.0  & 75.2  & 59.1  & 90.1  & 84.9  & 87.7  & 56.9         \\  
          EMI w/ APAA$_f$ ($\gamma$=0.8)        & 79.2  & 74.0  & 67.1  & 57.8  & 61.6  & 62.5  & 71.0  & 56.9  & 89.4  & 85.0  & 87.1  & 57.4         \\  
          EMI w/ APAA$_f$ ($\gamma$=1.0)        & 76.3  & 70.9  & 63.0  & 53.3  & 53.5  & 56.4  & 66.7  & 57.5  & 88.5  & 83.8  & 86.4  & 55.7         \\   \hline 
          VT~\cite{wang2021enhancing}           & 69.4  & 73.8  & 70.3  & 69.0  & 71.7  & 67.5  & 71.1  & 50.6  & 80.2  & 74.2  & 77.1  & 44.1         \\
          VT w/ APAA$_f$ ($\gamma$=0.2)         & 73.3  & 73.5  & 67.3  & 70.2  & 69.1  & 68.6  & 69.6  & 49.4  & 75.0  & 70.8  & 70.4  & 36.5         \\   
          VT w/ APAA$_f$ ($\gamma$=0.4)         & 79.5  & 79.1  & 74.9  & 74.1  & 75.9  & 71.3  & 74.4  & 53.7  & 81.2  & 73.9  & 77.5  & 42.9         \\    
          VT w/ APAA$_f$ ($\gamma$=0.6)         & 79.3  & 77.5  & 73.1  & 73.0  & 75.1  & 69.9  & 72.8  & 52.4  & 83.1  & 75.7  & 78.3  & 46.3         \\   
          \textbf{VT w/ APAA$_f$ ($\gamma$=0.8)}         & 79.2  & 77.5  & 72.3  & 70.1  & 74.7  & 68.7  & 72.3  & 52.5  & 83.6  & 76.2  & 79.9  & 46.6         \\      
          VT w/ APAA$_f$ ($\gamma$=1.0)         & 76.1  & 75.6  & 68.9  & 66.0  & 70.8  & 65.4  & 71.3  & 53.0  & 83.4  & 75.9  & 78.1  & 46.1         \\   \hline
          AIFGTM~\cite{zou2022making}           & 68.7  & 68.1  & 61.1  & 63.1  & 64.9  & 62.2  & 64.2  & 44.3  & 71.8  & 67.4  & 67.6  & 38.9         \\
          AIFGTM w/ APAA$_f$ ($\gamma$=0.2)     & 42.7  & 43.7  & 38.9  & 37.5  & 37.0  & 34.2  & 35.1  & 28.2  & 45.7  & 44.7  & 39.6  & 19.2         \\     
          AIFGTM w/ APAA$_f$ ($\gamma$=0.4)     & 64.8  & 65.4  & 59.3  & 60.4  & 60.5  & 57.3  & 59.4  & 42.0  & 68.0  & 63.8  & 62.7  & 33.4         \\  
          AIFGTM w/ APAA$_f$ ($\gamma$=0.6)     & 68.4  & 71.1  & 65.3  & 64.0  & 66.6  & 64.1  & 65.3  & 44.3  & 73.8  & 69.4  & 69.7  & 39.7         \\
          \textbf{AIFGTM w/ APAA$_f$ ($\gamma$=0.8)}     & 72.2  & 70.5  & 65.9  & 63.4  & 67.4  & 62.6  & 65.6  & 46.1  & 75.5  & 71.7  & 73.1  & 45.8         \\  
          AIFGTM w/ APAA$_f$ ($\gamma$=1.0)     & 72.8  & 69.3  & 65.4  & 63.3  & 67.1  & 62.3  & 65.4  & 46.3  & 74.7  & 71.9  & 73.9  & 46.3         \\   \hline
          \end{tabular}
        }
      \end{subtable}
    \end{center}
  \end{table*}

  \textbf{Evaluation Models.}
  We use both normally trained models and defense models to evaluate all attack methods. For CIFAR10, we use totally 8 normally trained models and 7 defense models for comprehensive evaluations. For ImageNet, we use totally 9 normally trained models and 12 defense models to evaluate. Specifically, the models used for CIFAR10 includes RegNet~\cite{radosavovic2020designing}, Res-18~\cite{he2016deep}, SENet-18~\cite{hu2018squeeze}, Dense-121~\cite{huang2017densely}, WideRes\textsubscript{28$\times$10}~\cite{zagoruyko2016wide}, DPN~\cite{chen2017dual}, Pyramid~\cite{han2017deep}, ShakeShake~\cite{gastaldi2017shake}, Dense-121\textsubscript{adv}~\cite{madry2018towards}, GoogLeNet\textsubscript{adv}~\cite{madry2018towards}, Res-18\textsubscript{adv}~\cite{madry2018towards}, $k$-WTA~\cite{xiao2020enhancing}, Odds~\cite{roth2019the}, Generative~\cite{li2019generative} and Ensemble~\cite{pang2019improving}.
  The models used for ImageNet include IncRes-v2~\cite{szegedy2017inception}, Inc-v3~\cite{szegedy2016rethinking}, Inc-v4~\cite{szegedy2017inception}, Res-101~\cite{he2016identity}, Res-152~\cite{he2016identity}, Mob\textsubscript{1.0}~\cite{sandler2018mobilenetv2}, Mob\textsubscript{1.4}~\cite{sandler2018mobilenetv2}, PNAS~\cite{liu2018progressive}, NAS~\cite{zoph2018learning}, Inc-v3\textsubscript{adv}~\cite{tramer2017ensemble}, Inc-v3\textsubscript{ens3}~\cite{tramer2017ensemble}, Inc-v3\textsubscript{ens4}~\cite{tramer2017ensemble}, IncRes-v2\textsubscript{ens}~\cite{tramer2017ensemble}, HGD~\cite{liao2018defense}, R\&P~\cite{xie2018mitigating}, NIPS-r3\footnote{\url{https://github.com/anlthms/nips-2017/tree/master/mmd}}, Bit-Red~\cite{xu2018feature}, JPEG~\cite{guo2018countering}, FD~\cite{liu2019feature}, ComDefend~\cite{jia2019comdefend} and RS~\cite{jia2020certified}.

  \textbf{Metrics.}
  We use the attack success rates on both white-box and black-box models to evaluation the effectiveness of different methods. Since all methods constrain the same $L_\infty$ perturbation budget, we additionally use mean absolute distance (MAD) and root mean square distance (RMSD) to compare the magnitude of perturbations generated by different methods.

  \begin{table*}[!t]
    \begin{center}
      \centering
      \caption{The attack success rates of the adversarial examples on \textbf{CIFAR10} under \textbf{untargeted attack} setting.}
      \label{tab:cifar_unt}
      \begin{subtable}[t]{\textwidth}
        \caption{The evaluation on the normally trained models.}
        \resizebox{\textwidth}{!}{
          \begin{tabular}{c|l|cccccccc|cc}
          \hline
          \multirow{2}{*}{\begin{tabular}[c]{@{}c@{}}Source\\ Model\end{tabular}} & \multicolumn{1}{c|}{\multirow{2}{*}{Method}} & \multicolumn{8}{c|}{Target Model}                                                          & \multicolumn{2}{c}{Distance Metric} \\ \cline{3-12} 
                                        & \multicolumn{1}{c|}{}                        & RegNet        & Res-18        & SENet-18       & Dense-121     & WideRes\textsubscript{28$\times$10}    & DPN    & Pyramid    & ShakeShake    & MAD                  & RMSD                 \\ \hline
          \multirow{6}{*}{RegNet}       & MIFGSM~\cite{dong2018boosting}                                 & 99.1          & 88.8          & 88.8           & 90.2          & 84.6          & 86.1          & 79.3          & 83.1          & 5.470          & 5.841          \\
                                        & MIFGSM w/ APAA$_f$                               & \textbf{99.8} & \textbf{92.7} & \textbf{92.3}  & \textbf{93.8} & \textbf{88.8} & \textbf{89.9} & \textbf{82.4} & \textbf{87.1} & \textbf{5.377} & \textbf{5.754} \\  \cline{2-12} 
                                        & DIM~\cite{xie2019improving}                                        & 98.0          & 91.0          & 90.7           & 91.8          & 88.2          & 88.2          & 85.1          & 86.6          & 5.544          & 5.892          \\
                                        & DIM w/ APAA$_f$                                   & \textbf{99.4} & \textbf{94.9} & \textbf{94.6}  & \textbf{95.5} & \textbf{92.5} & \textbf{92.3} & \textbf{89.2} & \textbf{91.3} & \textbf{5.309} & \textbf{5.682} \\               \cline{2-12} 
                                        & SIM~\cite{lin2020nesterov}                                         & 98.1          & 93.2          & 93.2           & 94.0          & 91.5          & 90.6          & 88.2          & 89.6          & 5.606          & 5.949          \\
                                        & SIM w/ APAA$_f$                                   & \textbf{99.4} & \textbf{96.5} & \textbf{96.4}  & \textbf{97.1} & \textbf{95.6} & \textbf{94.4} & \textbf{92.7} & \textbf{93.5} & \textbf{5.343} & \textbf{5.708} \\      \hline
          \multirow{6}{*}{\begin{tabular}[c]{@{}c@{}}RegNet\\+Res-18\\+SENet-18\\+Dense-121\end{tabular}}      & MIFGSM~\cite{dong2018boosting}                                      & 98.0          & 98.5          & 99.5           & 99.0          & 96.1          & 94.5          & 94.2          & 95.4          & 5.636          & 5.984          \\
                                        & MIFGSM w/ APAA$_f$                                & \textbf{99.4} & \textbf{99.6} & \textbf{100.0} & \textbf{99.9} & \textbf{98.7} & \textbf{97.8} & \textbf{97.5} & \textbf{98.2} & \textbf{5.496} & \textbf{5.856} \\                         \cline{2-12} 
                                        & DIM~\cite{xie2019improving}                                        & 97.6          & 98.4          & 99.3           & 98.7          & 96.6          & 94.7          & 94.8          & 95.9          & 5.688          & 6.017          \\
                                        & DIM w/ APAA$_f$                                   & \textbf{99.4} & \textbf{99.7} & \textbf{99.9}  & \textbf{99.7} & \textbf{98.7} & \textbf{97.6} & \textbf{97.9} & \textbf{98.5} & \textbf{5.427} & \textbf{5.783} \\               \cline{2-12} 
                                        & SIM~\cite{lin2020nesterov}                                         & 97.8          & 98.6          & 99.4           & 98.9          & 97.3          & 95.6          & 95.7          & 96.6          & 5.747          & 6.065          \\
                                        & SIM w/ APAA$_f$                                   & \textbf{99.3} & \textbf{99.7} & \textbf{99.9}  & \textbf{99.8} & \textbf{99.3} & \textbf{98.1} & \textbf{98.3} & \textbf{98.8} & \textbf{5.462} & \textbf{5.807}     \\              \hline
          \end{tabular}
        }
      \end{subtable}
      \begin{subtable}[t]{0.9\textwidth}
        \caption{The evaluation on the defense models.}
        \resizebox{\textwidth}{!}{
          \begin{tabular}{c|l|ccccccc}
          \hline
          \multirow{2}{*}{\begin{tabular}[c]{@{}c@{}}Source\\ Model\end{tabular}} & \multicolumn{1}{c|}{\multirow{2}{*}{Method}} & \multicolumn{7}{c}{Target Model}              \\ \cline{3-9} 
                                        & \multicolumn{1}{c|}{}                        & Dense-121\textsubscript{adv}   & GoogLeNet\textsubscript{adv}   & Res-18\textsubscript{adv}   & $k$-WTA   & Odds   & Generative   & Ensemble   \\ \hline
          \multirow{6}{*}{RegNet}       & MIFGSM~\cite{dong2018boosting}                                     & 21.0 & 45.1 & 15.9 & 75.5 & 87.0 & 53.6 & 79.7 \\
                                        & MIFGSM w/ APAA$_f$                                & \textbf{23.3} & \textbf{50.0} & \textbf{18.8} & \textbf{78.6} & \textbf{91.4} & \textbf{55.7} & \textbf{83.0} \\            \cline{2-9} 
                                        & DIM~\cite{xie2019improving}                                        & 23.4 & 53.5 & 17.9 & 80.1 & 90.8 & 59.9 & 84.2 \\
                                        & DIM w/ APAA$_f$                                   & \textbf{30.6} & \textbf{61.6} & \textbf{23.8} & \textbf{84.5} & \textbf{94.7} & \textbf{62.1} & \textbf{88.5} \\            \cline{2-9} 
                                        & SIM~\cite{lin2020nesterov}                                        & 25.0 & 57.5 & 18.8 & 84.5 & 93.6 & 63.0 & 87.9 \\
                                        & SIM w/ APAA$_f$                                   & \textbf{31.7} & \textbf{64.7} & \textbf{23.8} & \textbf{88.6} & \textbf{96.8} & \textbf{65.1} & \textbf{92.1} \\              \hline
          \multirow{6}{*}{\begin{tabular}[c]{@{}c@{}}RegNet\\+Res-18\\+SENet-18\\+Dense-121\end{tabular}}   & MIFGSM~\cite{dong2018boosting} & 34.6 & 69.5 & 25.6 & 92.1 & 97.5 & 72.2 & 93.2 \\
                                        & MIFGSM w/ APAA$_f$                               & \textbf{38.9} & \textbf{73.5} & \textbf{30.8} & \textbf{95.8} & \textbf{99.5} & \textbf{74.9} & \textbf{96.8} \\              \cline{2-9} 
                                        & DIM~\cite{xie2019improving}                                        & 36.4 & 71.8 & 27.2 & 92.7 & 98.4 & 74.1 & 94.1 \\
                                        & DIM w/ APAA$_f$                                  & \textbf{38.9} & \textbf{73.5} & \textbf{30.8} & \textbf{95.8} & \textbf{99.5} & \textbf{74.9} & \textbf{96.8} \\                \cline{2-9} 
                                        & SIM~\cite{lin2020nesterov}                                        & 39.0 & 73.9 & 28.4 & 94.0 & 99.0 & 75.4 & 95.2 \\
                                        & SIM w/ APAA$_f$                                   & \textbf{45.5} & \textbf{79.5} & \textbf{35.3} & \textbf{97.3} & \textbf{99.7} & \textbf{78.1} & \textbf{97.7} \\    \hline
          \end{tabular}
        }
      \end{subtable}
    \end{center}
  \end{table*}

  \textbf{Baselines.}
  We use MIFGSM~\cite{dong2018boosting}, DIM~\cite{xie2019improving}, TIM~\cite{dong2019evading}, SIM~\cite{lin2020nesterov}, VT~\cite{wang2021enhancing}, EMI~\cite{wang2021boosting}, AIFGTM~\cite{zou2022making}, SGM~\cite{wu2020skip}, IR~\cite{wang2021a} and SVRE~\cite{xiong2022stochastic} as baselines to compare with our proposed method. The details of the hyperparameters in these methods are provided in \cref{tab:baseline}. The number of iteration $T$ in the generation of adversarial examples for all methods, including ours, is 10 unless mentioned. We clip the adversarial examples to the range of the normal image (\ie, 0-255) and constrain the adversarial perturbations within the perturbation budget on $L_\infty$ norm bound in each step of the iteration process.

  \textbf{Details.}
  For the method of fixed scaling factor, we select 10000 training images of CIFAR10 and one subset of ImageNet\textsuperscript{\ref{foot:subset1}} as the validation set to search for the best scaling factor, and utilize the corresponding scaling factor to conduct the attack and evaluation in the testset of CIFAR10 and the other subset of ImageNet\textsuperscript{\ref{foot:subset2}}, respectively. By searching on the validation set, we set the scaling factor $\gamma$ as 8 for all methods on CIFAR10. For ImageNet, we use 0.4 in TIM, EMI, SGM, IR, and 0.8 in VT and AIFGTM as the scaling factor $\gamma$.
  We train the adaptive scaling factor generator with the training set of CIFAR10. The training of the model is quite fast. It takes about 2-3 hours on a GTX 1080Ti GPU with 3-4 epochs. The architecture of the generator is shown in \cref{tab:arch}.
  It should be mentioned that our proposed generator needs at least two white-box models to train. Through experiments, we find that when training the generator with only one classifier model, \ie, $n=1$ and $C_x=C_y$, the generated scaling factor is as high as possible. Although the higher white-box attack success rate can be achieved in this case, the transferability of generated adversarial examples is dropped dramatically. To maintain the transferability of adversarial examples, we use different classifier models when calculating the gradient information and updating the parameters of generator $G_t$, so that the generated scaling factor will not overfit to a specific model.

\subsection{Attack with Fixed Scaling Factor}
\label{sec:exp_fixed}
  We conduct comprehensive experiments on CIFAR10 and ImageNet to demonstrate the effectiveness of our proposed $\text{APAA}_f$, \ie, replacing the sign normalization with a fixed scaling factor. The value of the scaling factor in the experiments is determined by manually hyperparameter searching within a certain range on the validation set, which is totally separate from the test set used in the experiments below.

  \textbf{The influence of the scaling factor $\gamma$.}
  We analyze the influence of the hyperparameter $\gamma$ on the ImageNet dataset~\cite{russakovsky2015imagenet}. \cref{tab:imagenet_gamma} shows the attack success rates of various baseline methods combined with our proposed APAA, considering different values of $\gamma$ for normally trained and defense models. As $\gamma$ increases, the magnitude of the adversarial perturbation also increases, indicating a more aggressive perturbation. The attack success rate on normally trained models increases accordingly. However, for defense models, especially those obtained through adversarial training, the opposite conclusion is reached. Subtable (b) in \cref{tab:imagenet_gamma} reveals that as $\gamma$ increases, the attack success rate of adversarial examples on adversarially trained models gradually decreases. This behavior is attributed to models typically being trained with a fixed perturbation size during adversarial training. The model obtained in this manner exhibits better defense capabilities against moderate perturbations but reduced robustness when facing smaller adversarial perturbations, which are rarely seen during the training time. To strike a balance between achieving high attack success rates on normally trained models and defense models meanwhile minimizing the perturbation budgets introduced by our APAA method, we opt for moderate $\gamma$ coefficients for different methods: 0.4 for TIM and EMI, and 0.8 for VT and AIFGSM.

  \begin{table*}[!t]
    \begin{center}
      \centering
      \caption{The attack success rates of the adversarial examples on \textbf{ImageNet} under \textbf{untargeted attack} setting.}
      \label{tab:imagenet_unt}
      \begin{subtable}[t]{\textwidth}
        \caption{The evaluation on the normally trained models.}
        \resizebox{\textwidth}{!}{
          \begin{tabular}{c|l|ccccccccc|cc}
          \hline
          \multirow{2}{*}{\begin{tabular}[c]{@{}c@{}}Source\\ Model\end{tabular}} & \multicolumn{1}{c|}{\multirow{2}{*}{Method}} & \multicolumn{9}{c|}{Target Model}                                                                 & \multicolumn{2}{c}{Distance Metric} \\ \cline{3-13} 
                                                                                  & \multicolumn{1}{c|}{}                        & IncRes-v2     & Inc-v3         & Inc-v4        & Res-101       & Res-152 & Mob\textsubscript{1.0}    & Mob\textsubscript{1.4}    & PNAS    & NAS    & MAD              & RMSD              \\ \hline
          \multirow{8}{*}{IncRes-v2}                                              & TIM~\cite{dong2019evading}                                        & 98.3          & 86.5           & 83.9          & 73.3          & 73.8          & 81.4          & 83.7          & 75.1          & 77.3          & 10.302         & 11.108          \\
                                                                                  & TIM w/ APAA$_f$                                   & \textbf{98.6} & \textbf{88.6}  & \textbf{85.4} & \textbf{75.3} & \textbf{76.1} & \textbf{83.7} & \textbf{87.9} & \textbf{77.2} & \textbf{80.2} & \textbf{8.887} & \textbf{9.693}  \\                            \cline{2-13}
                                                                                  & VT~\cite{wang2021enhancing}                                          & 98.5          & 89.1           & 87.0          & 79.8          & 80.4          & 82.6          & 86.6          & 80.5          & 82.2          & 10.031         & 10.810          \\
                                                                                  & VT w/ APAA$_f$                                    & \textbf{99.5} & \textbf{92.3}  & \textbf{90.9} & \textbf{82.6} & \textbf{82.9} & \textbf{87.5} & \textbf{90.6} & \textbf{83.8} & \textbf{85.4} & \textbf{9.513} & \textbf{10.299} \\                             \cline{2-13} 
                                                                                  & EMI~\cite{wang2021boosting}                                        & 99.5          & 95.1           & 93.0          & 88.0          & 87.3          & 90.8          & 93.8          & 89.8          & 92.0          & 10.479         & 11.283          \\
                                                                                  & EMI w/ APAA$_f$                                   & \textbf{99.6} & \textbf{97.0}  & \textbf{95.9} & \textbf{90.2} & \textbf{90.5} & \textbf{93.6} & \textbf{95.2} & \textbf{91.8} & \textbf{92.7} & \textbf{9.390} & \textbf{10.234} \\                                         \cline{2-13} 
                                                                                  & AIFGTM~\cite{zou2022making}                                      & 97.4          & 82.2           & 78.0          & 72.5          & 72.4          & 77.1          & 80.4          & 74.0          & 75.1          & 10.415         & 11.269          \\
                                                                                  & AIFGTM w/ APAA$_f$                               & \textbf{98.3} & \textbf{84.2}  & \textbf{81.8} & \textbf{74.2} & \textbf{74.0} & \textbf{80.5} & \textbf{84.5} & \textbf{78.2} & \textbf{79.7} & \textbf{9.004} & \textbf{9.742}  \\                             \hline
          \multirow{8}{*}{\begin{tabular}[c]{@{}c@{}}IncRes-v2\\+Inc-v3\\+Inc-v4\\+Res-101\end{tabular}} & TIM~\cite{dong2019evading}                  & 98.5          & 99.4           & 99.0          & 97.0          & 93.0          & 92.8          & 94.3          & 91.8          & 93.5          & 10.289         & 11.155          \\
                                                                                  & TIM w/ APAA$_f$                                   & \textbf{99.7} & \textbf{99.9}  & \textbf{99.7} & \textbf{99.0} & \textbf{97.6} & \textbf{95.9} & \textbf{98.2} & \textbf{95.5} & \textbf{96.5} & \textbf{9.019} & \textbf{9.898}  \\                               \cline{2-13}
                                                                                  & VT~\cite{wang2021enhancing}                                          & 94.8          & 99.5           & 97.0          & 90.3          & 90.5          & 91.8          & 93.4          & 90.5          & 91.0          & 9.717          & 10.565          \\
                                                                                  & VT w/ APAA$_f$                                    & \textbf{98.3} & \textbf{99.7}  & \textbf{98.8} & \textbf{95.5} & \textbf{94.9} & \textbf{95.1} & \textbf{96.4} & \textbf{94.6} & \textbf{95.5} & \textbf{9.417} & \textbf{10.257} \\                                \cline{2-13} 
                                                                                  & EMI~\cite{wang2021boosting}                                         & 99.4          & 99.9           & 99.6          & 98.2          & 97.1          & 97.5          & 98.3          & 97.1          & 97.7          & 10.557         & 11.391          \\
                                                                                  & EMI w/ APAA$_f$                                   & \textbf{99.7} & \textbf{100.0} & \textbf{99.8} & \textbf{99.0} & \textbf{98.3} & \textbf{99.1} & \textbf{99.3} & \textbf{98.6} & \textbf{98.5} & \textbf{9.489} & \textbf{10.381} \\                                             \cline{2-13} 
                                                                                  & AIFGTM~\cite{zou2022making}                                      & 98.0          & 98.9           & 98.5          & 96.2          & 91.7          & 90.7          & 94.1          & 90.3          & 91.7          & 10.475         & 11.378          \\
                                                                                  & AIFGTM w/ APAA$_f$                                & \textbf{98.6} & \textbf{99.4}  & \textbf{99.1} & \textbf{97.6} & \textbf{94.5} & \textbf{93.7} & \textbf{96.0} & \textbf{93.5} & \textbf{94.0} & \textbf{9.070} & \textbf{9.865}                         \\ \hline
          \end{tabular}
        }
      \end{subtable}
      \begin{subtable}[t]{\textwidth}
        \caption{The evaluation on the defense models.}
        \resizebox{\textwidth}{!}{
          \begin{tabular}{c|l|cccccccccccc}
          \hline
          \multirow{2}{*}{\begin{tabular}[c]{@{}c@{}}Source\\ Model\end{tabular}} & \multicolumn{1}{c|}{\multirow{2}{*}{Method}} & \multicolumn{12}{c}{Target Model}                                                 \\ \cline{3-14} 
          & \multicolumn{1}{c|}{}                        & Inc-v3\textsubscript{adv} & Inc-v3\textsubscript{ens3} & Inc-v3\textsubscript{ens4} & IncRes-v2\textsubscript{ens} & HGD  & R\&P & NIPS-r3 & Bit-Red & JPEG & FD   & ComDefend & RS   \\ \hline
          \multirow{8}{*}{IncRes-v2}                                              & TIM~\cite{dong2019evading}                                         & 56.4          & 64.7          & 55.7          & 51.6          & 60.6          & 53.0          & 59.1          & 42.4          & 75.3          & 67.2          & 70.3          & 36.2          \\
                                                                                  & TIM w/ APAA$_f$                                   & \textbf{68.2} & \textbf{66.2} & \textbf{60.3} & \textbf{54.0} & \textbf{62.2} & \textbf{54.9} & \textbf{61.0} & \textbf{43.9} & \textbf{76.6} & \textbf{69.7} & \textbf{71.3} & \textbf{40.5} \\                                    \cline{2-14}
                                                                                  & VT~\cite{wang2021enhancing}                                          & 69.4          & 73.8          & 70.3          & 69.0          & 71.7          & 67.5          & 71.1          & 50.6          & 80.2          & 74.2          & 77.1          & 44.1          \\
                                                                                  & VT w/ APAA$_f$                                   & \textbf{79.2} & \textbf{77.5} & \textbf{72.3} & \textbf{70.1} & \textbf{74.7} & \textbf{68.7} & \textbf{72.3} & \textbf{52.5} & \textbf{83.6} & \textbf{76.2} & \textbf{79.9} & \textbf{46.6} \\                                            \cline{2-14} 
                                                                                  & EMI~\cite{wang2021boosting}                                        & 76.8          & 79.3          & 71.3          & 66.2          & 76.9          & 69.1          & 76.0          & 58.9          & 87.8          & 83.3          & 84.9          & 50.2          \\
                                                                                  & EMI w/ APAA$_f$                                  & \textbf{86.5} & \textbf{83.7} & \textbf{76.3} & \textbf{74.6} & \textbf{80.2} & \textbf{75.2} & \textbf{79.1} & \textbf{60.9} & \textbf{90.3} & \textbf{85.3} & \textbf{87.5} & \textbf{56.9} \\                                            \cline{2-14} 
                                                                                  & AIFGTM~\cite{zou2022making}                                     & 68.7 & 68.1 & 61.1 & 63.1 & 64.9 & 62.2 & 64.2 & 44.3 & 71.8 & 67.4 & 67.6 & 38.9 \\
                                                                                  & AIFGTM w/ APAA$_f$                                & \textbf{72.2} & \textbf{70.5} & \textbf{65.9} & \textbf{63.4} & \textbf{67.4} & \textbf{62.6} & \textbf{65.6} & \textbf{46.1} & \textbf{75.5} & \textbf{71.7} & \textbf{73.1} & \textbf{45.8} \\                                              \hline
          \multirow{8}{*}{\begin{tabular}[c]{@{}c@{}}IncRes-v2\\+Inc-v3\\+Inc-v4\\+Res-101\end{tabular}}   & TIM~\cite{dong2019evading}               & 85.2 & 86.7 & 85.3 & 78.3 & 86.6 & 79.9 & 84.3 & 62.9 & 91.4 & 83.6 & 88.4 & 56.2 \\
                                                                                  & TIM w/ APAA$_f$                                   & \textbf{92.8} & \textbf{91.1} & \textbf{90.0} & \textbf{83.0} & \textbf{91.5} & \textbf{85.7} & \textbf{89.0} & \textbf{65.0} & \textbf{94.1} & \textbf{87.5} & \textbf{90.9} & \textbf{61.1} \\                                                     \cline{2-14}
                                                                                  & VT~\cite{wang2021enhancing}                                         & 85.1 & 85.4 & 83.8 & 79.1 & 83.6 & 79.3 & 82.3 & 66.7 & 88.8 & 84.1 & 86.6 & 60.8 \\
                                                                                  & VT w/ APAA$_f$                                    & \textbf{93.6} & \textbf{93.1} & \textbf{91.5} & \textbf{86.0} & \textbf{91.6} & \textbf{87.2} & \textbf{89.6} & \textbf{73.4} & \textbf{95.0} & \textbf{88.4} & \textbf{92.7} & \textbf{69.0} \\                                                 \cline{2-14} 
                                                                                  & EMI~\cite{wang2021boosting}                                        & 92.5 & 91.4 & 89.5 & 84.4 & 91.2 & 85.6 & 89.6 & 75.6 & 95.9 & 91.6 & 94.3 & 68.8 \\
                                                                                  & EMI w/ APAA$_f$                                   & \textbf{96.6} & \textbf{95.6} & \textbf{93.9} & \textbf{89.4} & \textbf{95.1} & \textbf{91.0} & \textbf{92.7} & \textbf{78.8} & \textbf{97.2} & \textbf{93.6} & \textbf{95.5} & \textbf{73.0} \\                                                  \cline{2-14} 
                                                                                  & AIFGTM~\cite{zou2022making}                                     & 88.7 & 88.0 & 86.1 & 81.4 & 86.8 & 81.9 & 84.5 & 63.4 & 88.7 & 82.0 & 85.5 & 57.3 \\
                                                                                  & AIFGTM w/ APAA$_f$                               & \textbf{92.6} & \textbf{91.1} & \textbf{89.7} & \textbf{86.7} & \textbf{92.1} & \textbf{87.2} & \textbf{89.3} & \textbf{69.1} & \textbf{93.0} & \textbf{86.3} & \textbf{90.7} & \textbf{65.6}                                       \\ \hline
          \end{tabular}
        }
      \end{subtable}
    \end{center}
  \end{table*}

  \begin{table*}[!h]
    \begin{center}
      \centering
      \caption{The attack success rates of SGM~\cite{wu2020skip} and our APAA$_f$ method on \textbf{ImageNet} under \textbf{untargeted attack} setting.}
      \label{tab:imagenet_sgm}
      \begin{subtable}[t]{\textwidth}
        \caption{The evaluation on the normally trained models.}
        \resizebox{\textwidth}{!}{
          \begin{tabular}{c|l|ccccccccc|cc}
          \hline
          \multirow{2}{*}{\begin{tabular}[c]{@{}c@{}}Source\\ Model\end{tabular}} & \multicolumn{1}{c|}{\multirow{2}{*}{Method}} & \multicolumn{9}{c|}{Target Model}                                                                          & \multicolumn{2}{c}{Distance Metric} \\ \cline{3-13} 
                                                                                  & \multicolumn{1}{c|}{}                        & Dense-201      & Res-152       & Res-34  & VGG-16  & VGG-19  & SENet-154  & Inc-v3  & Inc-v4  & IncRes-v2  & MAD              & RMSD              \\ \hline
          \multirow{2}{*}{Dense-201}                                              & SGM~\cite{wu2020skip}                                        & \textbf{100.0} & 94.9          & 94.6    & 90.1    & 90.2    & 86.5       & 84.6    & 80.4    & 79.1       & 9.947            & 10.805           \\
                                                                                  & SGM w/ APAA$_f$                                   & \textbf{100.0} & \textbf{95.7} & \textbf{95.9} & \textbf{92.1} & \textbf{90.8} & \textbf{88.5} & \textbf{85.4} & \textbf{82.7} & \textbf{81.5} & \textbf{9.694} & \textbf{10.528} \\                       \hline
          \multirow{2}{*}{Res-152}                                                & SGM~\cite{wu2020skip}                                        & 87.4           & \textbf{99.9} & 92.7    & 89.9    & 88.1    & 77.6       & 79.6    & 74.3    & 72.9       & 10.089           & 10.988           \\
                                                                                  & SGM w/ APAA$_f$                                   & \textbf{87.8}  & \textbf{99.9} & \textbf{93.5} & \textbf{91.9} & \textbf{89.4} & \textbf{78.7} & \textbf{81.8} & \textbf{75.5} & \textbf{75.8} & \textbf{9.487} & \textbf{10.357} \\                     \hline
          \end{tabular}
        }
      \end{subtable}
      \begin{subtable}[t]{\textwidth}
        \caption{The evaluation on the defense models.}
        \resizebox{\textwidth}{!}{
          \begin{tabular}{c|l|cccccccccccc}
          \hline
          \multirow{2}{*}{\begin{tabular}[c]{@{}c@{}}Source\\ Model\end{tabular}} & \multicolumn{1}{c|}{\multirow{2}{*}{Method}} & \multicolumn{12}{c}{Target Model}                                                                                             \\ \cline{3-14} 
                                                                                  & \multicolumn{1}{c|}{}                        & Inc-v3\textsubscript{adv} & Inc-v3\textsubscript{ens3} & Inc-v3\textsubscript{ens4} & IncRes-v2\textsubscript{ens} & HGD  & R\&P & NIPS-r3 & Bit-Red & JPEG & FD   & ComDefend & RS   \\ \hline
          \multirow{2}{*}{Dense-201}                                              & SGM~\cite{wu2020skip}                                        & 71.8         & 66.6          & 67.4          & 58.2            & 79.4 & 58.1 & 64.6    & 48.6    & 71.2 & 65.6 & 58.0      & 53.0 \\
                                                                                  & SGM w/ APAA$_f$                                   & \textbf{75.6} & \textbf{70.1} & \textbf{69.8} & \textbf{60.4} & \textbf{82.7} & \textbf{62.4} & \textbf{67.4} & \textbf{51.3} & \textbf{73.2} & \textbf{66.3} & \textbf{60.1} & \textbf{54.7} \\                  \hline
          \multirow{2}{*}{Res-152}                                                & SGM~\cite{wu2020skip}                                        & 62.8         & 57.8          & 55.5          & 46.1            & 76.4 & 47.2 & 52.0    & 43.4    & 62.8 & 57.2 & 47.6      & 42.0 \\
                                                                                  & SGM w/ APAA$_f$                                   & \textbf{65.8} & \textbf{62.0} & \textbf{58.1} & \textbf{48.6} & \textbf{77.0} & \textbf{51.2} & \textbf{54.5} & \textbf{45.4} & \textbf{66.7} & \textbf{60.3} & \textbf{48.9} & \textbf{43.0}              \\ \hline
          \end{tabular}
        }
      \end{subtable}
    \end{center}
  \end{table*}

  \begin{table*}[!h]
    \begin{center}
      \centering
      \caption{The attack success rates of IR~\cite{wang2021a} and our APAA$_f$ method on \textbf{ImageNet} under \textbf{untargeted attack} setting. The results of the IR method are the results reported in their paper.}
      \label{tab:imagenet_ir}
      \resizebox{0.9\textwidth}{!}{
        \begin{tabular}{c|l|ccccccccc}
        \hline
        \multirow{2}{*}{\begin{tabular}[c]{@{}c@{}}Source\\ Model\end{tabular}} & \multicolumn{1}{c|}{\multirow{2}{*}{Method}} & \multicolumn{9}{c}{Target Model}                                                                              \\ \cline{3-11} 
                                                                                & \multicolumn{1}{c|}{}                        & Res-34        & Dense-121     & VGG-16  & Res-152  & Dense-201  & SENet-154  & Inc-v3  & Inc-v4  & IncRes-v2  \\ \hline
        \multirow{2}{*}{Res-34}                                                 & IR~\cite{wang2021a}                                         & -             & -             & 90.0          & 85.7          & 88.5          & 67.0          & 66.9          & 60.2          & 53.9          \\
                                                                                & IR w/ APAA$_f$                                    & \textbf{97.7} & \textbf{92.0} & \textbf{90.6} & \textbf{88.5} & \textbf{90.0} & \textbf{72.2} & \textbf{68.5} & \textbf{63.3} & \textbf{59.8} \\                                                \hline
        \multirow{2}{*}{Dense-121}                                              & IR~\cite{wang2021a}                                          & -             & -             & 89.0          & 83.2          & 93.4          & 74.2          & 69.6          & 64.7          & 58.2          \\
                                                                                & IR w/ APAA$_f$                                    & \textbf{89.8} & \textbf{98.1} & \textbf{89.7} & \textbf{87.7} & \textbf{96.2} & \textbf{78.9} & \textbf{73.3} & \textbf{70.0} & \textbf{66.5}           \\ \hline
        \end{tabular}
      }
    \end{center}
  \end{table*}
  \textbf{The untargeted attack.}
  The experiments of adversarial examples generated on CIFAR10 and ImageNet under the untargeted attack setting are shown in \cref{tab:cifar_unt} and \cref{tab:imagenet_unt} respectively. From the results on both the single model and the ensembled model attacks, combined with our proposed method of the scaling factor, the attack success rates of generated adversarial examples in all the state-of-the-art methods are improved. In addition, under the same perturbation budget constrain on $L_\infty$ norm, the MAD and RMSD between the original images and generated adversarial examples of our $\text{APAA}_f$ are smaller than the existing methods. Due to the accurate gradient directions used in our proposed method, our attack method is more effective, which has higher attack success rates with fewer perturbations. We also conduct experiments on SGM~\cite{wu2020skip} and IR~\cite{wang2021a} methods in \cref{tab:imagenet_sgm} and \cref{tab:imagenet_ir}, respectively. It demonstrates that our method can well integrate with almost all gradient-based attack methods to improve the attack success rates.
    

  \begin{table*}[!t]
    \begin{center}
      \centering
      \caption{The attack success rates of the adversarial examples on \textbf{CIFAR10} under \textbf{targeted attack} setting.}
      \label{tab:cifar_t}
      \resizebox{\textwidth}{!}{
        \begin{tabular}{c|l|cccccccc|cc}
        \hline
        \multirow{2}{*}{\begin{tabular}[c]{@{}c@{}}Source\\ Model\end{tabular}} & \multicolumn{1}{c|}{\multirow{2}{*}{Method}} & \multicolumn{8}{c|}{Target Model}                                                         & \multicolumn{2}{c}{Distance Metric} \\ \cline{3-12} 
                                      & \multicolumn{1}{c|}{}                        & RegNet        & Res-18        & SENet-18       & Dense-121     & WideRes\textsubscript{28$\times$10}    & DPN    & Pyramid    & ShakeShake    & MAD                  & RMSD                 \\ \hline
        \multirow{6}{*}{RegNet}       & MIFGSM~\cite{dong2018boosting}                                      & 82.6          & 44.7          & 44.1          & 46.3          & 39.9          & 42.8          & 36.7          & 40.2          & 5.363          & 5.734          \\
                                      & MIFGSM w/ APAA$_f$                               & \textbf{90.5} & \textbf{53.6} & \textbf{52.4} & \textbf{54.5} & \textbf{46.8} & \textbf{50.5} & \textbf{42.9} & \textbf{47.7} & \textbf{5.206} & \textbf{5.585} \\                \cline{2-12} 
                                      & DIM~\cite{xie2019improving}                                         & 71.2          & 45.4          & 45.0          & 46.5          & 41.5          & 43.0          & 39.7          & 41.9          & 5.436          & 5.790          \\
                                      & DIM w/ APAA$_f$                                   & \textbf{80.6} & \textbf{54.8} & \textbf{53.1} & \textbf{55.8} & \textbf{50.5} & \textbf{51.3} & \textbf{47.9} & \textbf{51.0} & \textbf{5.147} & \textbf{5.523} \\                \cline{2-12} 
                                      & SIM~\cite{lin2020nesterov}                                        & 75.0          & 50.6          & 50.2          & 52.4          & 48.1          & 48.1          & 44.8          & 47.7          & 5.488          & 5.837          \\
                                      & SIM w/ APAA$_f$                                   & \textbf{84.4} & \textbf{61.2} & \textbf{60.0} & \textbf{62.5} & \textbf{58.2} & \textbf{57.2} & \textbf{54.1} & \textbf{56.8} & \textbf{5.168} & \textbf{5.537} \\             \hline
        \multirow{6}{*}{\begin{tabular}[c]{@{}c@{}}RegNet\\+Res-18\\+SENet-18\\+Dense-121\end{tabular}}      & MIFGSM~\cite{dong2018boosting}               & 77.8          & 82.1          & 87.6          & 85.7          & 70.2          & 65.0          & 65.6          & 69.9          & 5.513          & 5.865          \\
                                      & MIFGSM w/ APAA$_f$                                & \textbf{89.0} & \textbf{91.9} & \textbf{94.9} & \textbf{93.7} & \textbf{82.8} & \textbf{76.6} & \textbf{76.8} & \textbf{82.2} & \textbf{5.365} & \textbf{5.729} \\                    \cline{2-12} 
                                      & DIM~\cite{xie2019improving}                                         & 73.5          & 78.8          & 83.9          & 80.8          & 69.3          & 62.8          & 65.3          & 69.3          & 5.552          & 5.889          \\
                                      & DIM w/ APAA$_f$                                   & \textbf{84.8} & \textbf{88.8} & \textbf{92.0} & \textbf{90.1} & \textbf{81.1} & \textbf{74.4} & \textbf{76.7} & \textbf{80.8} & \textbf{5.289} & \textbf{5.650} \\                \cline{2-12} 
                                      & SIM~\cite{lin2020nesterov}                                        & 75.8          & 81.6          & 86.2          & 83.7          & 73.1          & 65.7          & 68.5          & 72.7          & 5.582          & 5.911          \\
                                      & SIM w/ APAA$_f$                                  & \textbf{87.1} & \textbf{91.0} & \textbf{93.8} & \textbf{92.3} & \textbf{84.7} & \textbf{77.2} & \textbf{79.5} & \textbf{83.5} & \textbf{5.279} & \textbf{5.632}                  \\               \hline
        \end{tabular}
      }
    \end{center}
  \end{table*}
  \textbf{The targeted attack.} 
  The experiments of adversarial examples generated on CIFAR10 under the targeted attack setting are shown in \cref{tab:cifar_t}. The target label of each image is randomly chosen among the 9 wrong labels. The average black-box attack success rates of the adversarial examples generated by $\text{APAA}_f$ against white-box and black-box models are about 10\% higher than those of baselines under both settings of the single model attack and the ensembled model attack. It verifies that our proposed scaling factor method is also effective under the targeted attack setting.

  \textbf{The influence of the size of perturbation.}
  \begin{figure*}[!t]
    \centering
    \begin{subfigure}[t]{0.31\textwidth}
      \centering
      \includegraphics[width=\textwidth]{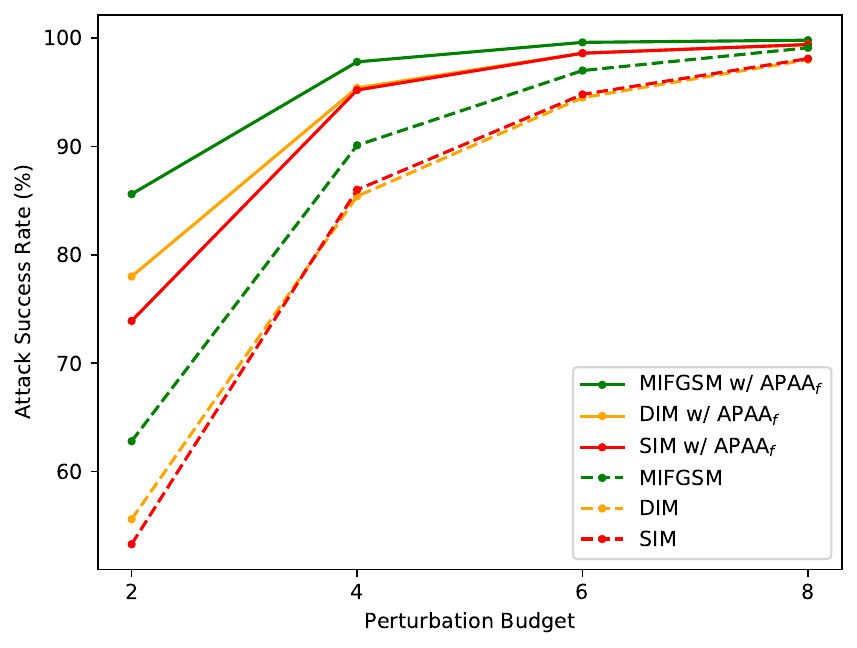}
      \caption{White-box model}
      \end{subfigure}
      \quad
      \begin{subfigure}[t]{0.31\textwidth}
      \centering
      \includegraphics[width=\textwidth]{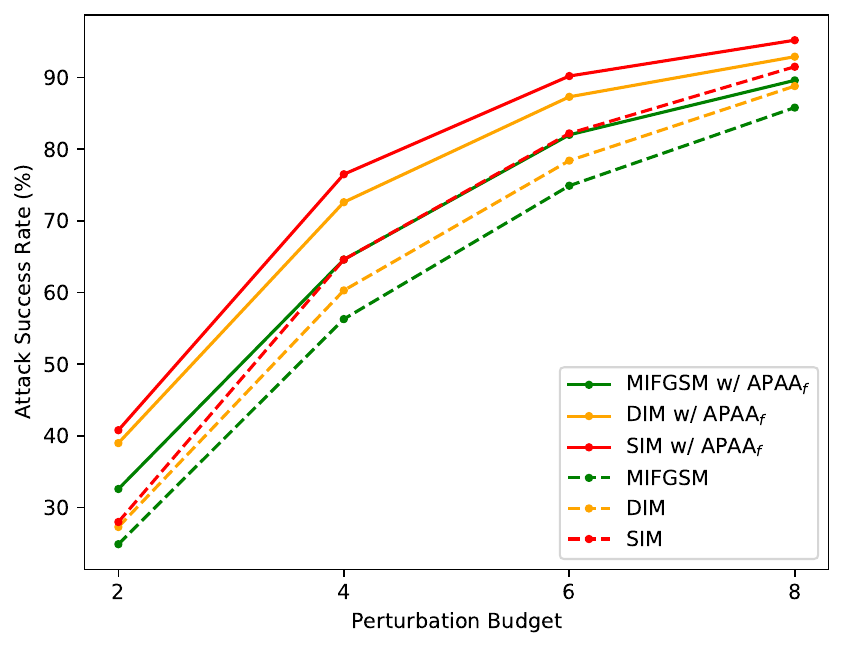}
      \caption{Black-box normally trained model}
    \end{subfigure}
    \quad
    \begin{subfigure}[t]{0.31\textwidth}
      \centering
      \includegraphics[width=\textwidth]{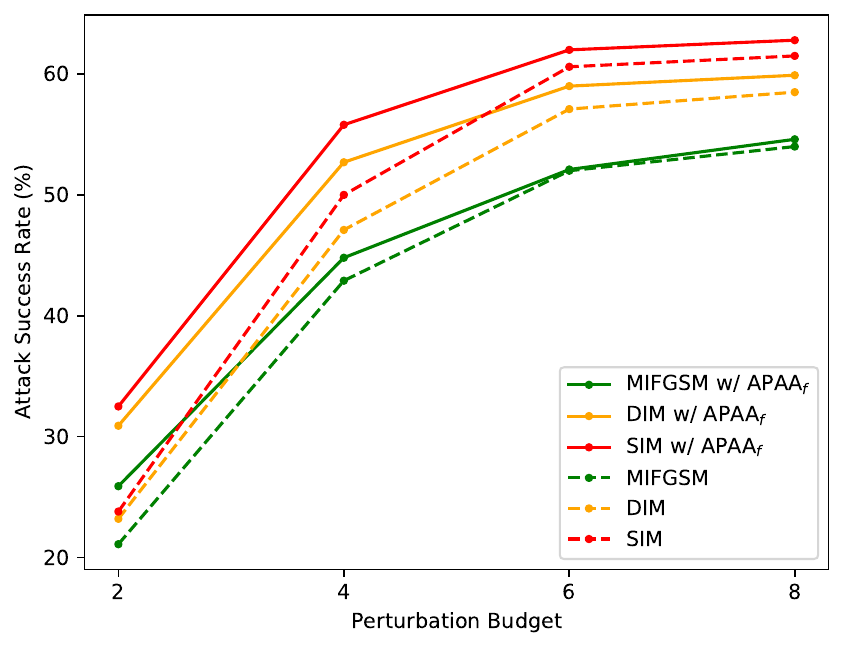}
      \caption{Defense model}
    \end{subfigure}
    \caption{The attack success rates vs. perturbation budget curve on CIFAR10. The curves with dotted lines are the results of baseline methods, and those with solid lines are the results of our method. Three subfigures are the average attack success rates of different methods on the white-box model, the black-box normally trained models and the defense models, respectively. The experiment chooses RegNet as the white-box model.}
    \label{fig:perturbation}
  \end{figure*}
  We conduct experiments to demonstrate the attack success rates vs. perturbation budget curves on CIFAR10 in \cref{fig:perturbation}. We can clearly find that our proposed method of using the scaling factor improves the attack success rates on various perturbation budgets, which shows the excellent generalization of our method.

\begin{table*}[!t]
  \begin{center}
    \centering
    \caption{The attack success rates of the adversarial examples on \textbf{CIFAR10} under untargeted attack setting with \textbf{different number of iterations}.}
    \label{tab:cifar_iter}
    \resizebox{\textwidth}{!}{
      \begin{tabular}{c|l|c|ccccccccccc}
      \hline
      \multirow{2}{*}{\begin{tabular}[c]{@{}c@{}}Source\\ Model\end{tabular}}      & \multicolumn{1}{c|}{\multirow{2}{*}{Method}} & \multirow{2}{*}{\begin{tabular}[c]{@{}c@{}}Step\\ Number\end{tabular}} & \multicolumn{11}{c}{Target Model}      \\ \cline{4-14} 
      & \multicolumn{1}{c|}{}      &   & RegNet & Res-18        & SENet-18 & Dense-121     & DPN  & \multicolumn{1}{c|}{ShakeShake} & Dense\textsubscript{adv} & GoogLeNet\textsubscript{adv} & Res-18\textsubscript{adv} & $k$-WTA & Ensemble \\ \hline
      \multirow{24}{*}{\begin{tabular}[c]{@{}c@{}}Res-18\\ +Dense-121\end{tabular}}
      & \multirow{4}{*}{MIFGSM~\cite{dong2018boosting}} & 8 & 92.5 & 98.5 & 95.8 & 99.4 & 89.9 & \multicolumn{1}{c|}{92.4} & 27.6 & 58.1 & 22.0 & 86.9 & 89.0       \\
      & & 10 & 92.9 & 98.7 & 96.1 & 99.6 & 90.3 & \multicolumn{1}{c|}{92.8} & 28.7 & 60.0 & 22.4 & 87.2 & 89.5       \\      
      & & 100 & 94.4 & 99.1 & 97.0 & 99.8 & 92.0 & \multicolumn{1}{c|}{94.3} & 32.0 & 63.5 & 22.7 & 88.6 & 90.9      \\
      & & 300 & 94.5 & 99.1 & 97.1 & 99.8 & 92.2 & \multicolumn{1}{c|}{94.4} & 32.0 & 63.6 & 22.7 & 88.6 & 91.0      \\ \cline{2-14} 
      & \multirow{4}{*}{MIFGSM w/ APAA$_f$} & 8 & 95.3  & 99.4  & 98.0  & 99.8  & 92.6  & \multicolumn{1}{c|}{95.5}  & 29.0  & 59.0  & 24.4  & 89.9  & 92.1       \\
      & & 10 & 96.3  & 99.7  & 98.4  & 99.9  & 93.5  & \multicolumn{1}{c|}{96.4}  & 32.6  & 62.7  & 26.1  & 90.9  & 92.8       \\      
      & & 100 & 97.1  & 100.0  & 99.0  & 100.0  & 94.0  & \multicolumn{1}{c|}{96.9}  & 35.8  & 66.3  & 27.9  & 90.0  & 92.4    \\
      & & 300 & 97.0  & 99.9  & 99.0  & 100.0  & 94.1  & \multicolumn{1}{c|}{96.8}  & 36.6  & 67.0  & 28.1  & 91.0  & 93.3     \\ \cline{2-14}
      & \multirow{4}{*}{DIM~\cite{xie2019improving}} & 8 & 92.7  & 98.3  & 95.9  & 98.7  & 90.0  & \multicolumn{1}{c|}{93.0}  & 27.4  & 60.6  & 22.7  & 87.5  & 89.8       \\
      & & 10 & 93.3  & 98.6  & 96.6  & 99.0  & 90.7  & \multicolumn{1}{c|}{93.8}  & 28.8  & 62.9  & 23.4  & 88.4  & 90.6       \\      
      & & 100 & 96.3  & 99.4  & 98.1  & 99.6  & 94.4  & \multicolumn{1}{c|}{96.2}  & 32.6  & 68.3  & 24.9  & 91.9  & 93.8      \\
      & & 300 & 96.6  & 99.4  & 98.2  & 99.7  & 94.8  & \multicolumn{1}{c|}{96.7}  & 32.2  & 68.5  & 24.5  & 92.3  & 94.1      \\ \cline{2-14}
      & \multirow{4}{*}{DIM w/ APAA$_f$} & 8 & 95.8  & 99.2  & 97.9  & 99.3  & 93.1  & \multicolumn{1}{c|}{95.8}  & 30.0  & 62.5  & 26.3  & 91.4  & 93.2       \\
      & & 10 & 96.8  & 99.6  & 98.5  & 99.7  & 94.5  & \multicolumn{1}{c|}{97.0}  & 35.7  & 67.9  & 29.2  & 91.2  & 93.2       \\      
      & & 100 & 98.9  & 99.9  & 99.5  & 100.0  & 96.9  & \multicolumn{1}{c|}{98.7}  & 36.6  & 69.9  & 27.1  & 94.6  & 96.4     \\
      & & 300 & 99.1  & 100.0  & 99.7  & 100.0  & 96.3  & \multicolumn{1}{c|}{98.8}  & 36.3  & 70.6  & 29.1  & 95.1  & 96.6    \\ \cline{2-14}
      & \multirow{4}{*}{SIM~\cite{lin2020nesterov}} & 8 & 94.0  & 98.4  & 96.8  & 98.9  & 91.7  & \multicolumn{1}{c|}{94.2}  & 20.5  & 53.8  & 19.9  & 90.4  & 92.1       \\
      & & 10 & 96.4  & 99.3  & 98.2  & 99.5  & 94.4  & \multicolumn{1}{c|}{96.3}  & 23.9  & 55.9  & 20.7  & 92.8  & 94.4       \\      
      & & 100 & 96.5  & 99.4  & 98.4  & 99.7  & 95.1  & \multicolumn{1}{c|}{96.7}  & 34.8  & 70.6  & 26.5  & 93.2  & 94.6      \\
      & & 300 & 96.8  & 99.5  & 98.5  & 99.7  & 95.3  & \multicolumn{1}{c|}{96.9}  & 32.5  & 67.2  & 26.4  & 93.4  & 94.9      \\ \cline{2-14}
      & \multirow{4}{*}{SIM w/ APAA$_f$} & 8 & 96.7  & 99.5  & 98.7  & 99.7  & 94.5  & \multicolumn{1}{c|}{97.0}  & 31.5  & 64.7  & 26.8  & 93.1  & 94.6       \\
      & & 10 & 97.9  & 99.7  & 99.2  & 99.9  & 95.6  & \multicolumn{1}{c|}{97.8}  & 32.5  & 65.3  & 27.2  & 93.7  & 95.3       \\      
      & & 100 & 98.9  & 100.0  & 99.7  & 100.0  & 97.1  & \multicolumn{1}{c|}{98.8}  & 35.5  & 70.0  & 26.8  & 95.4  & 96.6    \\
      & & 300 & 99.0  & 100.0  & 99.8  & 100.0  & 97.2  & \multicolumn{1}{c|}{98.9}  & 34.5  & 70.8  & 28.9  & 95.6  & 96.8    \\ \hline
      \end{tabular}
    }
  \end{center}
\end{table*}

\textbf{The influence of the number of attack steps.}
  We conduct experiments to compare the results of different methods under various settings of attack iteration steps. From \cref{tab:cifar_iter}, it is obvious that as the number of attack steps increases, the success rates of all attack methods improve. However, when comparing attack methods with the same number of steps, our $\text{APAA}_f$ consistently achieves higher attack success rates than baselines. Furthermore, when comparing the results of our 10-step $\text{APAA}_f$ with the 100-step baseline methods, it can be observed that our $\text{APAA}_f$ achieves comparable or even higher attack success rates with fewer attack steps. It confirms our opinion that accurate gradient directions can conduct a successful attack with fewer attack steps. This observation well validates the idea presented in~\cref{fig:motivation}.

  \begin{table*}[!t]
    \begin{center}
      \centering
      \caption{Compare with other alternatives to the sign function (\ie, arctanh and adam).}
      \label{tab:cifar_adam}
      \resizebox{\textwidth}{!}{
        \begin{tabular}{c|l|ccccccccccc}
        \hline
        \multirow{2}{*}{\begin{tabular}[c]{@{}c@{}}Source\\ Model\end{tabular}}      & \multicolumn{1}{c|}{\multirow{2}{*}{Method}} & \multicolumn{11}{c}{Target Model}                                                                                                                         \\ \cline{3-13} 
        & \multicolumn{1}{c|}{}                        & RegNet & Res-18        & SENet-18 & Dense-121     & DPN  & \multicolumn{1}{c|}{ShakeShake} & Dense\textsubscript{adv} & GoogLeNet\textsubscript{adv} & Res-18\textsubscript{adv} & $k$-WTA & Ensemble \\ \hline
        \multirow{12}{*}{\begin{tabular}[c]{@{}c@{}}RegNet\end{tabular}} 
        & MIFGSM~\cite{dong2018boosting}    & 99.1  & 88.8  & 88.8  & 90.2  & 86.1  & \multicolumn{1}{c|}{83.1}  & 21.0  & 45.1  & 15.9  & 75.5  & 79.7       \\
        & MIFGSM w/ Arctanh                 & 84.0  & 55.8  & 57.4  & 58.7  & 55.5  & \multicolumn{1}{c|}{48.9}  & 18.3  & 47.3  & 10.6  & 43.7  & 49.5       \\                  
        & MIFGSM w/ Adam                    & \textbf{100.0}  & 92.2  & 91.5  & 93.0  & 87.8  & \multicolumn{1}{c|}{85.0}  & 6.6  & 10.0  & 7.4  & 76.5  & 79.6       \\                  
        & MIFGSM w/ APAA$_f$                & 99.8  & \textbf{92.7}  & \textbf{92.3}  & \textbf{93.8}  & \textbf{89.9}  & \multicolumn{1}{c|}{\textbf{87.1}}  & \textbf{23.3}  & \textbf{50.0}  & \textbf{18.8}  & \textbf{78.6}  & \textbf{83.0}       \\   \cline{2-13} 
        & DIM~\cite{xie2019improving}       & 98.0  & 91.0  & 90.7  & 91.8  & 88.2  & \multicolumn{1}{c|}{86.6}  & 23.4  & 53.5  & 17.9  & 80.1  & 84.2       \\                  
        & DIM w/ Arctanh                    & 75.4  & 54.1  & 55.5  & 56.4  & 52.7  & \multicolumn{1}{c|}{47.5}  & 28.7  & 48.3  & 22.1  & 42.8  & 49.7       \\                  
        & DIM w/ Adam                       & 98.8  & 92.5  & 92.4  & 93.1  & 90.1  & \multicolumn{1}{c|}{88.4}  & 28.2  & 53.1  & 20.9  & 80.6  & 85.8       \\                  
        & DIM w/ APAA$_f$                   & \textbf{99.4}  & \textbf{94.9}  & \textbf{94.6}  & \textbf{95.5}  & \textbf{92.3}  & \multicolumn{1}{c|}{\textbf{91.3}}  & \textbf{30.6}  & \textbf{61.6}  & \textbf{23.8}  & \textbf{84.5}  & \textbf{88.5}       \\    \cline{2-13} 
        & SIM~\cite{lin2020nesterov}        & 98.1  & 93.2  & 93.2  & 94.0  & 90.6  & \multicolumn{1}{c|}{89.6}  & 25.0  & 57.5  & 18.8  & 84.5  & 87.9       \\
        & SIM w/ Arctanh                    & 79.3  & 61.1  & 62.4  & 63.6  & 58.6  & \multicolumn{1}{c|}{53.9}  & 24.0  & 53.5  & 23.4  & 48.0  & 55.9       \\
        & SIM w/ Adam                       & 98.0  & 90.4  & 91.0  & 92.0  & 86.2  & \multicolumn{1}{c|}{84.9}  & 20.7  & 55.8  & 18.6  & 78.0  & 83.5       \\
        & SIM w/ APAA$_f$                   & \textbf{99.4}  & \textbf{96.5}  & \textbf{96.4}  & \textbf{97.1}  & \textbf{94.4}  & \multicolumn{1}{c|}{\textbf{93.5}}  & \textbf{31.7}  & \textbf{64.7}  & \textbf{23.8}  & \textbf{88.6}  & \textbf{92.1}       \\ \hline
        \end{tabular}
      }
    \end{center}
  \end{table*}
  
\begin{table*}[!hbtp]
  \begin{center}
    \centering
    \caption{Compare with several generative methods.}
    \label{tab:cifar_generative}
    \resizebox{\textwidth}{!}{
      \begin{tabular}{c|l|cccccccccc|cc}
      \hline
      \multirow{2}{*}{\begin{tabular}[c]{@{}c@{}}Source\\ Model\end{tabular}} & \multicolumn{1}{c|}{\multirow{2}{*}{Method}} & \multicolumn{10}{c|}{Target Model}  & \multicolumn{2}{c}{Distance Metric} \\ \cline{3-14} 
      & \multicolumn{1}{c|}{}                        & Res-18    & SENet-18 & Dense-121      & DPN  & \multicolumn{1}{c|}{ShakeShake} & Dense-121\textsubscript{adv}    & GoogLeNet\textsubscript{adv} & Res-18\textsubscript{adv} & $k$-WTA & Ensemble & MAD              & RMSD              \\ \hline
      \multirow{4}{*}{RegNet}
      & AdvGAN~\cite{xiao2018generating}            & 83.0 & 86.4 & 82.2 & 66.0 & \multicolumn{1}{c|}{73.0} & 7.6 & 14.5 & 6.5 & 66.5 & 72.1 & 6.021 & 6.268       \\
      & NaturalAdversary~\cite{zhao2018generating}  & 80.2 & 79.5 & 82.4 & 77.2 & \multicolumn{1}{c|}{79.9} & \textbf{47.7} & 60.9 & \textbf{40.3} & 78.3 & 79.4 & 13.512 & 17.238      \\
      & SIM~\cite{lin2020nesterov}                  & 93.2 & 93.2 & 94.0 & 90.6 & \multicolumn{1}{c|}{89.6} & 25.0 & 57.5 & 18.8 & 84.5 & 87.9 & 5.606 & 5.949       \\
      & SIM w/ APAA$_f$                             & \textbf{96.5} & \textbf{96.4} & \textbf{97.1} & \textbf{94.4} & \multicolumn{1}{c|}{\textbf{93.5}} & 31.7 & \textbf{64.7} & 23.8 & \textbf{88.6} & \textbf{92.1} & \textbf{5.343} & \textbf{5.708}       \\  \hline
      \end{tabular}
    }
  \end{center}
\end{table*}

\textbf{Comparison with other alternative methods to the sign function.}
  To provide a more comprehensive demonstration of the superiority of our approach, which employs a scaling factor to replace the sign function method, we integrat the baseline methods with both the Adam optimizer~\cite{kingma2015adam} and the arctanh function, subsequently comparing them with our proposed $\text{APAA}_f$. The results presented in~\cref{tab:cifar_adam} demonstrate that the method utilizing the arctanh function yields unsatisfactory performance, which is even worse than the baselines. On the other hand, the utilization of the Adam optimizer, while showing some improvements over baseline methods, still falls short of matching the performance of our $\text{APAA}_f$. Notably, when combined with SIM, our APAA achieves  average improvements of 5\%-10\% in attack success rate, which demonstrates the superiority of APAA. We suppose that the Adam optimizer, originally designed for model parameter optimization, may not be also suitable for generating adversarial examples, which could potentially result in suboptimal results. Similarly, the arctanh function may not be well-suited for the task of generating adversarial examples.

  \begin{table*}[!hbtp]
    \begin{center}
      \centering
      \caption{The comparisons between the attack success rates of the adversarial examples by baselines, our proposed method with \textbf{fixed scaling factor} (APAA$_f$) and our proposed method with \textbf{adaptive scaling factor} (APAA$_a$) against both normally trained models and defense models on \textbf{CIFAR10} under untargeted attack setting.}
      \label{tab:cifar_adaptive}
      \resizebox{\textwidth}{!}{
        \begin{tabular}{c|l|cccccccccc|cc}
        \hline
        \multirow{2}{*}{\begin{tabular}[c]{@{}c@{}}Source\\ Model\end{tabular}} & \multicolumn{1}{c|}{\multirow{2}{*}{Method}} & \multicolumn{10}{c|}{Target Model}  & \multicolumn{2}{c}{Distance Metric} \\ \cline{3-14} 
        & \multicolumn{1}{c|}{}                        & RegNet & Res-18        & SENet-18      & DPN  & \multicolumn{1}{c|}{ShakeShake} & Dense-121\textsubscript{adv}    & GoogLeNet\textsubscript{adv} & Res-18\textsubscript{adv} & $k$-WTA & Ensemble & MAD              & RMSD              \\ \hline
        \multirow{12}{*}{\begin{tabular}[c]{@{}c@{}}Res-18\\+Dense-121\textsubscript{adv}\end{tabular}}
        & MIFGSM~\cite{dong2018boosting}    & 84.4 & 95.5 & 89.3 & 81.0 & \multicolumn{1}{c|}{84.8} & \textbf{94.8} & 57.4 & 21.3 & 78.6 & 80.9 & 5.349 & 5.763 \\
        & MIFGSM w/ APAA$_f$                 & 85.6 & 97.0 & 90.8 & 81.5 & \multicolumn{1}{c|}{86.2} & 94.7 & 63.1 & 23.0 & 79.2 & 82.2 & 4.976 & 5.390 \\   
        & MIFGSM w/ APAA$_a$                 & \textbf{86.5} & \textbf{97.3} & \textbf{91.8} & \textbf{82.6} & \multicolumn{1}{c|}{\textbf{87.4}} & \textbf{94.8} & \textbf{64.4} & \textbf{27.6} & \textbf{80.5} & \textbf{83.1} & \textbf{4.488} & \textbf{5.304} \\   \cline{2-14}
        & DIM~\cite{xie2019improving}        & 84.9 & 95.5 & 90.2 & 82.0 & \multicolumn{1}{c|}{86.1} & 86.2 & 57.2 & 22.1 & 80.1 & 82.4 & 5.428 & 5.824 \\
        & DIM w/ APAA$_f$                    & 86.6 & 96.7 & 91.8 & 82.4 & \multicolumn{1}{c|}{87.3} & 86.7 & 65.6 & 23.1 & 80.4 & 83.3 & 4.892 & 5.292 \\
        & DIM w/ APAA$_a$                    & \textbf{87.7} & \textbf{97.7} & \textbf{92.9} & \textbf{84.0} & \multicolumn{1}{c|}{\textbf{88.8}} & \textbf{87.4} & \textbf{66.1} & \textbf{29.9} & \textbf{82.4} & \textbf{85.3} & \textbf{4.208} & \textbf{5.049} \\   \cline{2-14}
        & SIM~\cite{lin2020nesterov}         & 89.7 & 97.3 & 93.7 & 86.4 & \multicolumn{1}{c|}{90.4} & 85.2 & 68.1 & 25.3 & 85.3 & 87.6 & 5.565 & 5.943 \\
        & SIM w/ APAA$_f$                    & 91.6 & \textbf{99.1} & 95.7 & 88.3 & \multicolumn{1}{c|}{92.4} & 87.5 & 70.8 & 26.0 & 86.9 & 89.9 & 5.034 & 5.416 \\
        & SIM w/ APAA$_a$                    & \textbf{93.8} & \textbf{98.5} & \textbf{96.7} & \textbf{90.3} & \multicolumn{1}{c|}{\textbf{94.2}} & \textbf{88.1} & \textbf{74.5} & \textbf{34.2} & \textbf{89.1} & \textbf{91.5} & \textbf{4.284} & \textbf{5.151} \\   \cline{2-14}
        & SVRE~\cite{xiong2022stochastic}    & 93.0 & 99.2 & 97.0 & 91.2 & \multicolumn{1}{c|}{94.5} & 85.5 & 64.5 & 19.7 & 87.6 & 90.2 & 5.564 & 5.957 \\
        & SVRE w/ APAA$_f$                   & 94.9 & 99.5 & 97.6 & 92.4 & \multicolumn{1}{c|}{95.2} & 87.8 & 72.3 & 22.8 & 89.7 & 92.8 & 4.915 & 5.322 \\
        & SVRE w/ APAA$_a$                   & \textbf{95.2} & \textbf{99.7} & \textbf{97.9} & \textbf{92.9} & \multicolumn{1}{c|}{\textbf{95.7}} & \textbf{88.6} & \textbf{73.8} & \textbf{25.6} & \textbf{90.3} & \textbf{93.4} & \textbf{4.903} & \textbf{5.309} \\    \hline
        \multirow{12}{*}{\begin{tabular}[c]{@{}c@{}}SENet-18\\+Dense-121\textsubscript{adv}\end{tabular}}
        & MIFGSM~\cite{dong2018boosting}    & 87.7 & 92.2 & 99.3 & 85.0 & \multicolumn{1}{c|}{88.9} & 94.4 & 62.3 & 22.8 & 82.8 & 85.7 & 5.426 & 5.832 \\
        & MIFGSM w/ APAA$_f$                 & 89.6 & 93.8 & 99.6 & \textbf{86.7} & \multicolumn{1}{c|}{\textbf{90.9}} & \textbf{95.2} & \textbf{68.2} & 24.2 & 84.3 & 87.3 & 5.071 & 5.478 \\   
        & MIFGSM w/ APAA$_a$                 & \textbf{90.0} & \textbf{94.0} & \textbf{99.7} & 86.4 & \multicolumn{1}{c|}{90.1} & \textbf{95.2} & 68.0 & \textbf{28.5} & \textbf{84.8} & \textbf{87.9} & \textbf{4.667} & \textbf{5.457} \\   \cline{2-14}
        & DIM~\cite{xie2019improving}        & 89.0 & 93.0 & 98.9 & 86.2 & \multicolumn{1}{c|}{90.0} & 88.1 & 64.1 & 23.6 & 84.5 & 87.3 & 5.496 & 5.883 \\
        & DIM w/ APAA$_f$                    & 90.4 & 94.2 & \textbf{99.6} & 87.4 & \multicolumn{1}{c|}{\textbf{92.2}} & 88.5 & 70.5 & 25.0 & 85.4 & 89.0 & 5.018 & 5.415 \\
        & DIM w/ APAA$_a$                    & \textbf{91.6} & \textbf{95.2} & 99.3 & \textbf{88.7} & \multicolumn{1}{c|}{92.0} & \textbf{89.7} & \textbf{73.6} & \textbf{32.6} & \textbf{86.8} & \textbf{90.1} & \textbf{4.259} & \textbf{5.126} \\   \cline{2-14}
        & SIM~\cite{lin2020nesterov}         & 92.0 & 95.1 & 99.2 & 89.7 & \multicolumn{1}{c|}{92.7} & 87.6 & 73.4 & 26.7 & 88.4 & 90.9 & 5.613 & 5.981 \\
        & SIM w/ APAA$_f$                    & 95.2 & 97.6 & \textbf{99.8} & \textbf{93.1} & \multicolumn{1}{c|}{95.5} & 89.6 & 75.1 & 27.4 & \textbf{91.8} & \textbf{94.2} & 5.160 & 5.539 \\
        & SIM w/ APAA$_a$                    & \textbf{95.3} & \textbf{97.7} & \textbf{99.8} & \textbf{93.1} & \multicolumn{1}{c|}{\textbf{95.7}} & \textbf{89.8} & \textbf{75.4} & \textbf{31.3} & 91.4 & 94.1 & \textbf{4.871} & \textbf{5.661} \\  \cline{2-14}
        & SVRE~\cite{xiong2022stochastic}    & 94.9 & 97.1 & 99.6 & 92.5 & \multicolumn{1}{c|}{95.3} & 85.9 & 67.5 & 19.7 & 91.4 & 93.6 & 5.680 & 6.062 \\
        & SVRE w/ APAA$_f$                   & 96.3 & 98.5 & \textbf{99.8} & 93.6 & \multicolumn{1}{c|}{96.8} & 87.0 & \textbf{73.5} & 23.0 & 92.5 & 95.2 & 5.089 & 5.494 \\
        & SVRE w/ APAA$_a$                   & \textbf{96.7} & \textbf{98.9} & \textbf{99.8} & \textbf{93.9} & \multicolumn{1}{c|}{\textbf{97.1}} & \textbf{87.6} & 73.1 & \textbf{24.1} & \textbf{92.9} & \textbf{95.6} & \textbf{4.913} & \textbf{5.236} \\    \hline
              \multirow{12}{*}{\begin{tabular}[c]{@{}c@{}}Dense-121\\+Res-18\textsubscript{adv}\end{tabular}}
      & MIFGSM~\cite{dong2018boosting}    & 87.4 & 90.5 & 91.1 & 84.5 & \multicolumn{1}{c|}{85.3} & 44.8 & 65.9 & 83.2 & 80.3 & 82.3 & 5.414 & 5.811 \\
      & MIFGSM w/ APAA$_f$                 & 91.1 & 94.1 & 94.1 & 88.0 & \multicolumn{1}{c|}{88.9} & 48.6 & 68.6 & 89.9 & 84.2 & 86.3 & 5.264 & 5.655 \\   
      & MIFGSM w/ APAA$_a$                 & \textbf{91.8} & \textbf{94.5} & \textbf{94.5} & \textbf{88.6} & \multicolumn{1}{c|}{\textbf{89.2}} & \textbf{49.5} & \textbf{69.3} & \textbf{91.2} & \textbf{84.6} & \textbf{86.8} & \textbf{5.159} & \textbf{5.597} \\   \cline{2-14}
      & DIM~\cite{xie2019improving}        & 87.3 & 90.2 & 90.7 & 84.8 & \multicolumn{1}{c|}{85.8} & 46.0 & 67.9 & 79.7 & 80.5 & 83.2 & 5.476 & 5.859 \\
      & DIM w/ APAA$_f$                    & 91.7 & 94.1 & 94.5 & 89.2 & \multicolumn{1}{c|}{90.3} & 50.5 & 70.3 & 85.4 & 85.2 & 87.8 & 5.150 & 5.535 \\
      & DIM w/ APAA$_a$                    & \textbf{92.5} & \textbf{94.8} & \textbf{95.1} & \textbf{89.8} & \multicolumn{1}{c|}{\textbf{91.0}} & \textbf{51.5} & \textbf{71.2} & \textbf{86.1} & \textbf{86.1} & \textbf{88.6} & \textbf{5.047} & \textbf{5.446} \\   \cline{2-14}
      & SIM~\cite{lin2020nesterov}         & 90.7 & 93.3 & 93.7 & 88.7 & \multicolumn{1}{c|}{90.1} & 52.2 & 74.8 & 77.9 & 86.4 & 88.3 & 5.610 & 5.971 \\
      & SIM w/ APAA$_f$                    & 94.8 & 96.7 & 96.9 & 92.8 & \multicolumn{1}{c|}{94.2} & 55.1 & 76.6 & 86.7 & 90.5 & 92.8 & \textbf{5.250} & \textbf{5.616} \\
      & SIM w/ APAA$_a$                    & \textbf{95.5} & \textbf{97.2} & \textbf{97.4} & \textbf{93.5} & \multicolumn{1}{c|}{\textbf{94.8}} & \textbf{56.6} & \textbf{79.1} & \textbf{87.2} & \textbf{91.5} & \textbf{92.9} & 5.559 & 6.222 \\   \cline{2-14}
      & SVRE~\cite{xiong2022stochastic}    & 91.9 & 94.8 & 94.5 & 90.4 & \multicolumn{1}{c|}{91.2} & 39.8 & 66.6 & 81.8 & 87.7 & 88.8 & 5.721 & 6.079 \\
      & SVRE w/ APAA$_f$                   & 94.3 & 96.4 & 95.8 & 93.0 & \multicolumn{1}{c|}{93.6} & 45.7 & 71.9 & 83.9 & 89.8 & 90.3 & 5.047 & 5.449 \\
      & SVRE w/ APAA$_a$                   & \textbf{94.8} & \textbf{96.7} & \textbf{96.0} & \textbf{93.7} & \multicolumn{1}{c|}{\textbf{93.9}} & \textbf{48.9} & \textbf{73.0} & \textbf{84.6} & \textbf{90.4} & \textbf{91.7} & \textbf{4.976} & \textbf{5.276} \\    \hline
        \multirow{12}{*}{\begin{tabular}[c]{@{}c@{}}Res-18\\+SENet-18\\+Dense-121\textsubscript{adv}\\+GoogLeNet\textsubscript{adv}\end{tabular}}
        & MIFGSM~\cite{dong2018boosting}    & 93.2 & 97.1 & 98.7 & 91.0 & \multicolumn{1}{c|}{93.6} & 93.2 & 99.7 & 33.7 & 89.4 & 91.4 & 5.337 & 5.761 \\
        & MIFGSM w/ APAA$_f$                 & 95.6 & 98.8 & 99.5 & 93.7 & \multicolumn{1}{c|}{96.1} & \textbf{96.1} & \textbf{99.9} & 37.0 & \textbf{92.2} & \textbf{94.2} & 5.061 & 5.471 \\   
        & MIFGSM w/ APAA$_a$                 & \textbf{95.9} & \textbf{99.0} & \textbf{99.6} & \textbf{94.0} & \multicolumn{1}{c|}{\textbf{96.2}} & 96.0 & \textbf{99.9} & \textbf{38.6} & 92.1 & \textbf{94.2} & \textbf{4.947} & \textbf{5.712} \\   \cline{2-14}
        & DIM~\cite{xie2019improving}        & 93.9 & 97.2 & 98.6 & 91.8 & \multicolumn{1}{c|}{94.0} & 88.6 & 97.8 & 34.8 & 90.0 & 92.1 & 5.461 & 5.858 \\
        & DIM w/ APAA$_f$                    & 96.1 & 98.7 & 99.3 & 94.2 & \multicolumn{1}{c|}{96.2} & \textbf{91.4} & \textbf{99.3} & 37.5 & 92.8 & 94.7 & \textbf{5.050} & \textbf{5.449} \\
        & DIM w/ APAA$_a$                    & \textbf{96.5} & \textbf{99.0} & \textbf{99.5} & \textbf{94.5} & \multicolumn{1}{c|}{\textbf{96.7}} & \textbf{91.4} & \textbf{99.3} & \textbf{37.7} & \textbf{93.2} & \textbf{95.2} & 5.075 & 5.814 \\   \cline{2-14}
        & SIM~\cite{lin2020nesterov}         & 96.9 & 98.8 & 99.5 & 95.5 & \multicolumn{1}{c|}{97.1} & 89.7 & 95.8 & 38.8 & 94.6 & 96.1 & 5.891 & 6.477 \\
        & SIM w/ APAA$_f$                    & \textbf{98.1} & \textbf{99.5} & \textbf{99.8} & \textbf{96.9} & \multicolumn{1}{c|}{98.2} & \textbf{92.5} & 97.8 & 41.5 & \textbf{96.1} & \textbf{97.5} & 5.202 & 5.942 \\
        & SIM w/ APAA$_a$                    & 98.0 & \textbf{99.5} & \textbf{99.8} & 96.8 & \multicolumn{1}{c|}{\textbf{98.4}} & \textbf{92.5} & \textbf{97.9} & \textbf{47.1} & 95.8 & 97.1 & \textbf{4.896} & \textbf{5.687} \\   \cline{2-14}
        & SVRE~\cite{xiong2022stochastic}    & 97.5 & 99.4 & 99.6 & 95.8 & \multicolumn{1}{c|}{97.8} & 88.4 & 96.5 & 30.8 & 95.4 & 96.6 & 5.642 & 6.031 \\
        & SVRE w/ APAA$_f$                   & 98.5 & 99.8 & \textbf{99.9} & 97.3 & \multicolumn{1}{c|}{98.8} & 90.4 & 98.5 & 35.6 & 96.6 & 97.5 & 5.159 & 5.554 \\
        & SVRE w/ APAA$_a$                   & \textbf{98.7} & \textbf{99.9} & \textbf{99.9} & \textbf{97.5} & \multicolumn{1}{c|}{\textbf{99.1}} & \textbf{90.7} & \textbf{98.8} & \textbf{36.0} & \textbf{96.8} & \textbf{97.9} & \textbf{4.817} & \textbf{5.233} \\    \hline
        \end{tabular}
      }
    \end{center}
  \end{table*}

  \begin{figure}[t]
    \centering
    \includegraphics[width=\columnwidth]{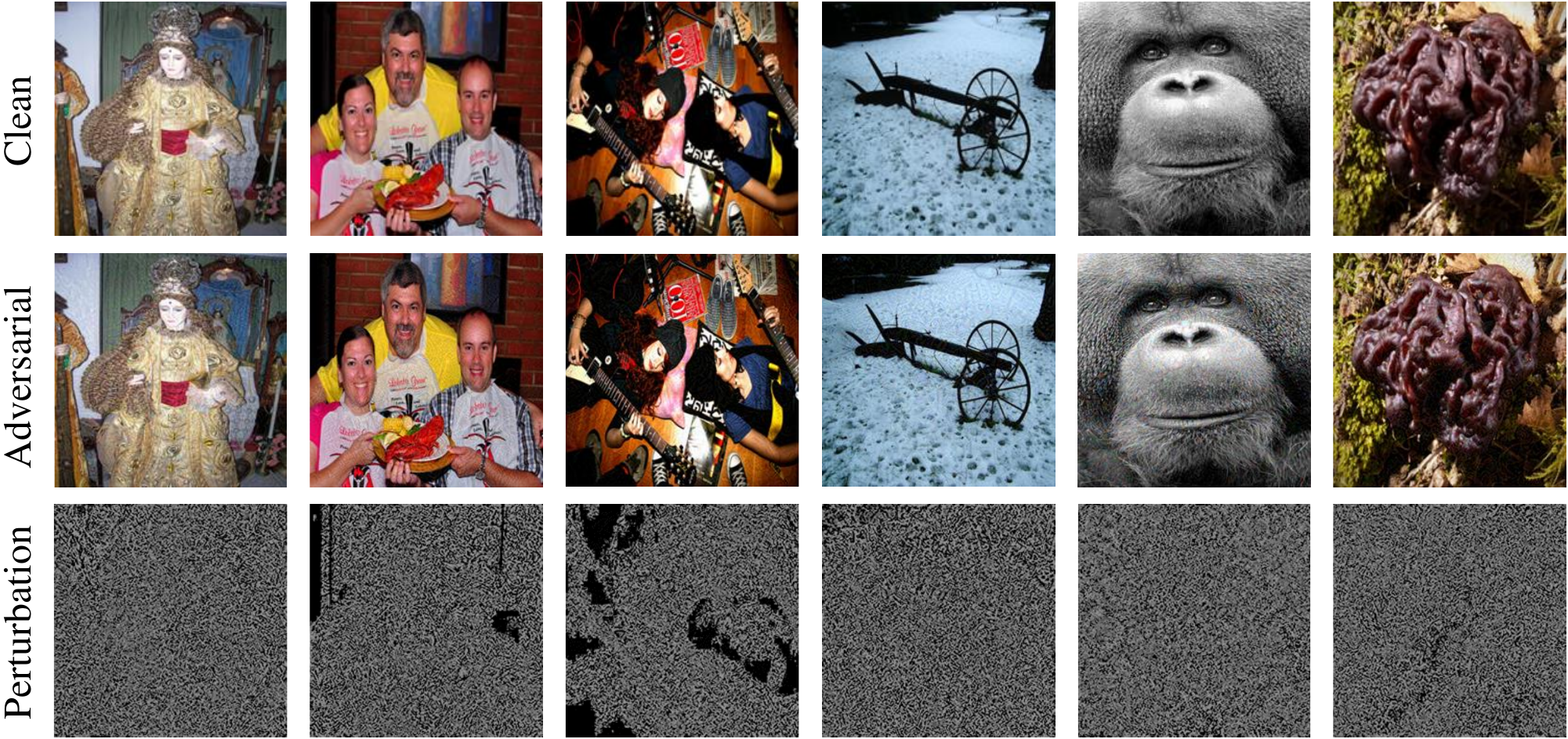}
    \caption{Visualization of adversarial examples crafted on IncRes-v2 by MIFGSM~\cite{dong2018boosting} with our proposed APAA$_f$.}
    \label{fig:visualization}
  \end{figure}

  \textbf{Comparison with generative attack methods.}
  We further compare our proposed APAA with some typical generative attack methods~\cite{xiao2018generating, zhao2018generating}. \cref{tab:cifar_generative} demonstrates that in comparison to AdvGAN~\cite{xiao2018generating}, which shares consistent experimental settings with our study, our $\text{APAA}_f$ method consistently achieves significantly higher attack success rates with smaller adversarial perturbations.  
  In contrast to NaturalAdversary~\cite{zhao2018generating}, which conducting attack without perturbation constraints, thus introducing much larger perturbations compared to APAA, we still attain higher attack success rates across nearly all models (except for Dense-121$\textsubscript{adv}$). Therefore, our APAA method also exhibits clear advantages over generative attack methods. We suppose that the poor performance of generative attack methods may be attributed to the fact that generative adversarial attack methods typically operate in a black-box fashion. Attackers may struggle to precisely control the generated adversarial examples, resulting in the poor transferability on black-box models.

  \textbf{The visualization of adversarial examples.}
  We provide a visualization of our generated adversarial examples. As shown in \cref{fig:visualization}, since our method can conduct the attack with relatively fewer perturbations, the generated adversarial perturbations are almost imperceptible.

\subsection{Attack with Adaptive Scaling Factor}
\label{sec:exp_adaptive}

  We further demonstrate the effectiveness of our proposed adaptive scaling factor generator. The process of generating adversarial examples with $\text{APAA}_a$ is summarized in \cref{alg:infer}. For fair comparisons, we consider all the models used during the training of the generator ($C_1, \cdots, C_n$) as white-box models, and conduct both attacks of SIM and our method on the ensemble of all white-box models.

  To conduct a comprehensive comparison among our proposed $\text{APAA}_f$, $\text{APAA}_a$ methods, and the baseline methods, \cref{tab:cifar_adaptive} presents a comparison of our $\text{APAA}_f$ and $\text{APAA}_a$ methods when combined with MIFGSM, DIM, SIM and SVRE, respectively, against baseline methods under six different source model settings. From \cref{tab:cifar_adaptive}, it becomes apparent that in most cases, $\text{APAA}_a$ achieves higher attack success rates than $\text{APAA}_f$. However, in some instances, $\text{APAA}_f$ uses smaller perturbations than $\text{APAA}_a$. Nevertheless, both $\text{APAA}_a$ and $\text{APAA}_f$ consistently outperform baseline methods in terms of attack success rates across various settings.
  It is noteworthy that although SVRE is a stronger baseline compared to MIFGSM, DIM and SIM, after integration with our APAA, SVRE achieves higher attack success rates with fewer perturbations. This further demonstrates the effectiveness of our APAA method again.
  Due to the distinct advantages of these two methods, attackers can freely select one of them based on the specific scenario when conducting the attack. For experiments on more models, please refer to \cref{tab:more_cifar_adaptive} in the appendix.

  \begin{algorithm}[t]
    \algnewcommand\algorithmicinput{\textbf{Input:}}
    \algnewcommand\Input{\item[\algorithmicinput]}
    \algnewcommand\algorithmicoutput{\textbf{Output:}}
    \algnewcommand\Output{\item[\algorithmicoutput]}
  
    \caption{Generating adversarial examples with adaptive scaling factor generator}
    \label{alg:infer}
    \begin{algorithmic}[1]
        \Input the original image $x$ and corresponding label $y$
        \Input the number of attack iteration $T$
        \Input the scaling factor generators $G_1, G_2, \cdots, G_T$
        \Input the ensembled model $C$, which contains all white-box classifier models $C_1, \cdots, C_n$
        \Output the adversarial example $\bm{x_T^{adv}}$
        \State $\bm{x_0^{adv}} = \bm{x}$
        \For {t $\in \{1, \cdots T\}$ }
          \State $\bm{grad_t}=\nabla_{\bm{x}} J(C(\bm{x_{t-1}^{adv}}), y)$
          \State $\gamma_t=G_t(\bm{x_{t-1}^{adv}}, \bm{grad_t})$
          \State $\bm{x_t^{adv}}= \Pi_{\bm{x}, \epsilon}(\gamma_t \cdot \bm{grad_t} + \bm{x_{t-1}^{adv}})$ \Comment{using the update in BIM as an example} 
        \EndFor
        \State \Return $\bm{x_T^{adv}}$
    \end{algorithmic}
  \end{algorithm}

\section{Conclusion}
\label{sec:conclusion}
  In this work, we propose to use the scaling factor instead of the sign function to normalize the gradient of the input example when conducting the adversarial attack, which can achieve a more accurate gradient direction and thus improve the attack success rate. The scaling factor can either be elaborately selected manually, or adaptively achieved by a generator according to different image characteristics. We also theoretically demonstrate that our proposed method can improve the black-box transferability of adversarial examples. Extensive experiments on CIFAR10 and ImageNet show the superiority of our proposed methods, which can improve the black-box attack success rates on both normally trained models and defense models with fewer update steps and perturbation budgets.



\ifCLASSOPTIONcaptionsoff
  \newpage
\fi



%



\bibliographystyle{IEEEtran}
\bibliography{egbib}

%
\begin{IEEEbiography}[{\includegraphics[width=1in,height=1.25in,clip,keepaspectratio]{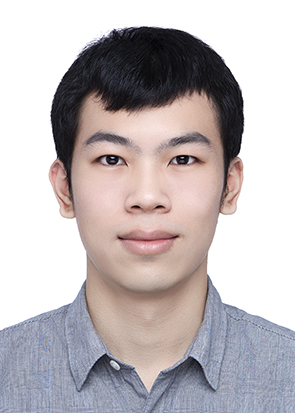}}]{Zheng Yuan}
  received the B.S. degree from University of Chinese Academy of Sciences in 2019. He is currently pursuing the Ph.D. degree from University of Chinese Academy of Sciences. His research interest includes adversarial example and model robustness. He has authored several academic papers in international conferences including ICCV/ECCV/ICPR.

\end{IEEEbiography}

\begin{IEEEbiography}[{\includegraphics[width=1in,height=1.25in,clip,keepaspectratio]{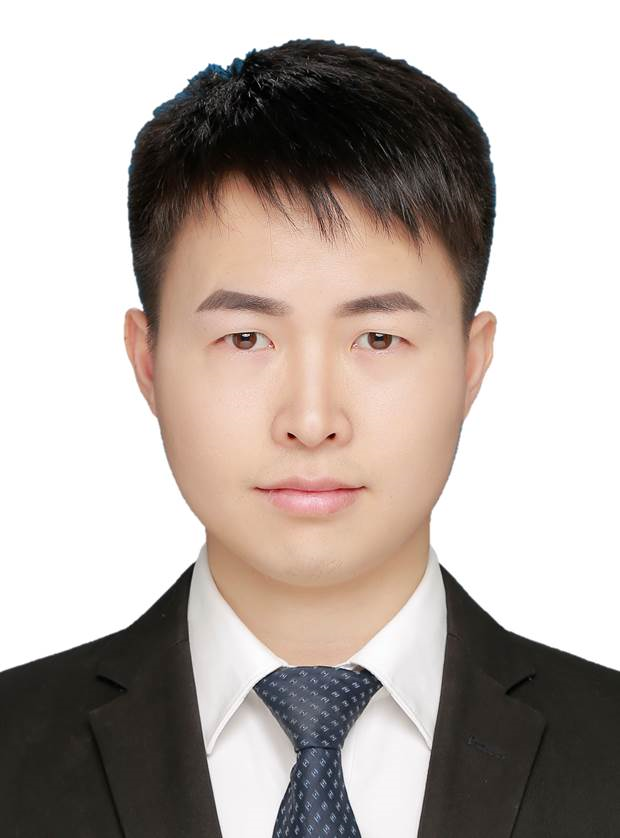}}]{Jie Zhang}
  is an associate professor with the Institute of Computing Technology, Chinese Academy of Sciences (CAS). He received the Ph.D. degree from the University of Chinese Academy of Sciences, Beijing, China. His research interests cover computer vision, pattern recognition, machine learning, particularly include face recognition, image segmentation, weakly/semi-supervised learning, domain generalization.
\end{IEEEbiography}


\begin{IEEEbiography}[{\includegraphics[width=1in,height=1.25in,clip,keepaspectratio]{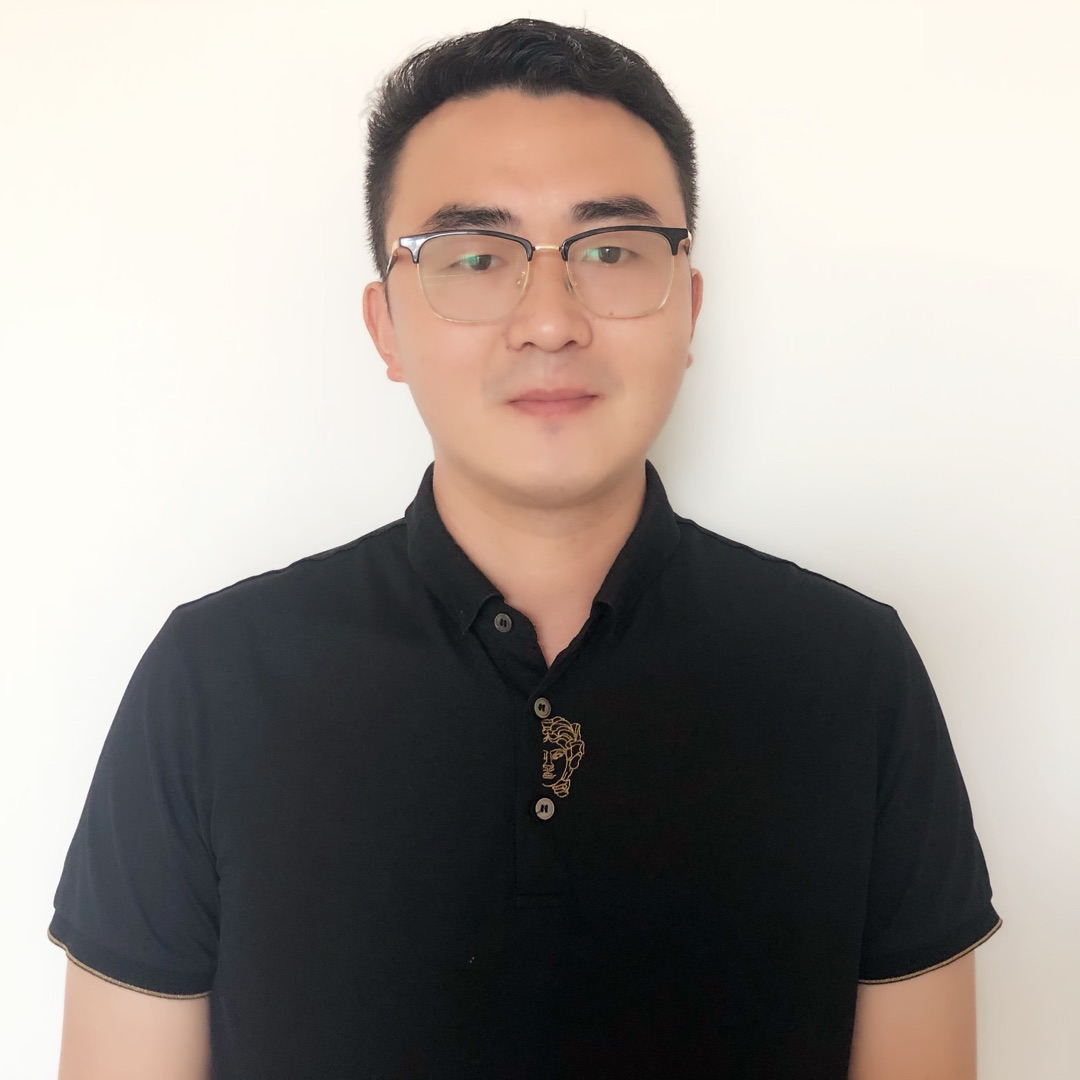}}]{Zhaoyan Jiang}
  received the  MS degree in Pattern Recognition and Intelligent System from Northeastern University in 2013. He currently works as a research engineer in Tencent, Inc, Beijing, China. His research interests include computer vision, metric learning and image processing.
\end{IEEEbiography}

\begin{IEEEbiography}[{\includegraphics[width=1in,height=1.25in,clip,keepaspectratio]{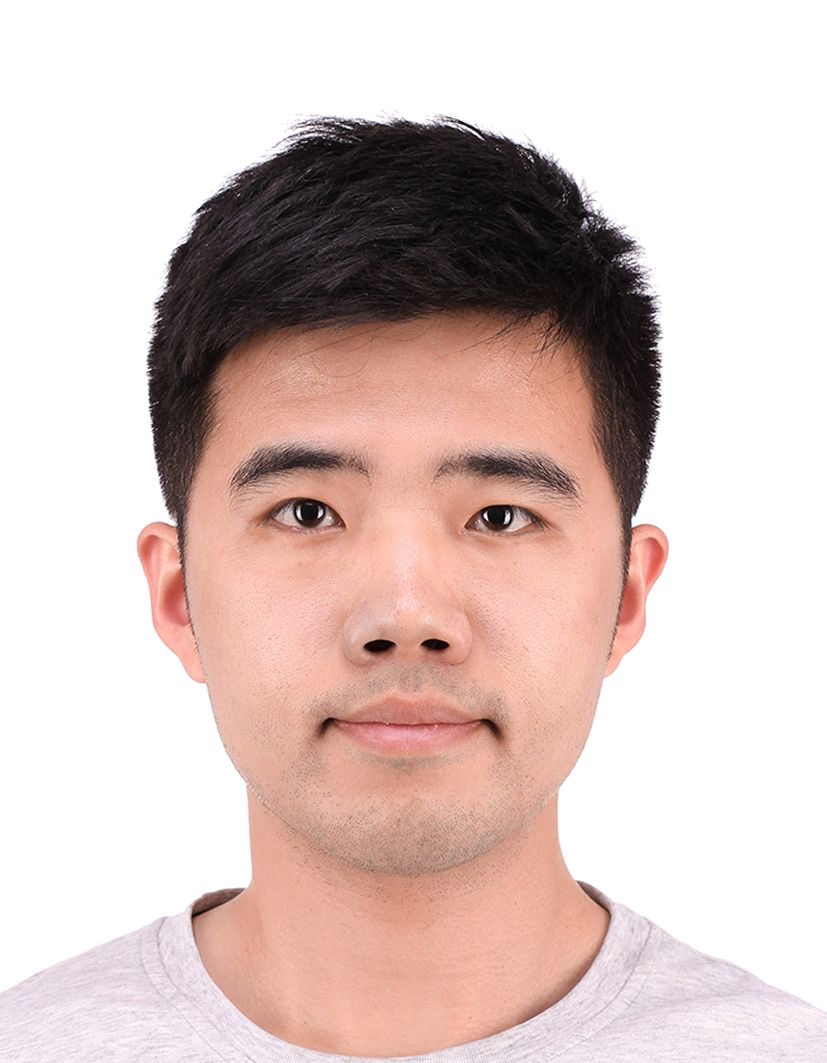}}]{Liangliang Li}
  received the  MS degree in  Computer Science and Technology from  Tsinghua University in 2018. He currently works as a research engineer in Tencent, Inc, Beijing, China. His research interests include computer vision, metric learning and image processing.
\end{IEEEbiography}

\begin{IEEEbiography}[{\includegraphics[width=1in,height=1.25in,clip,keepaspectratio]{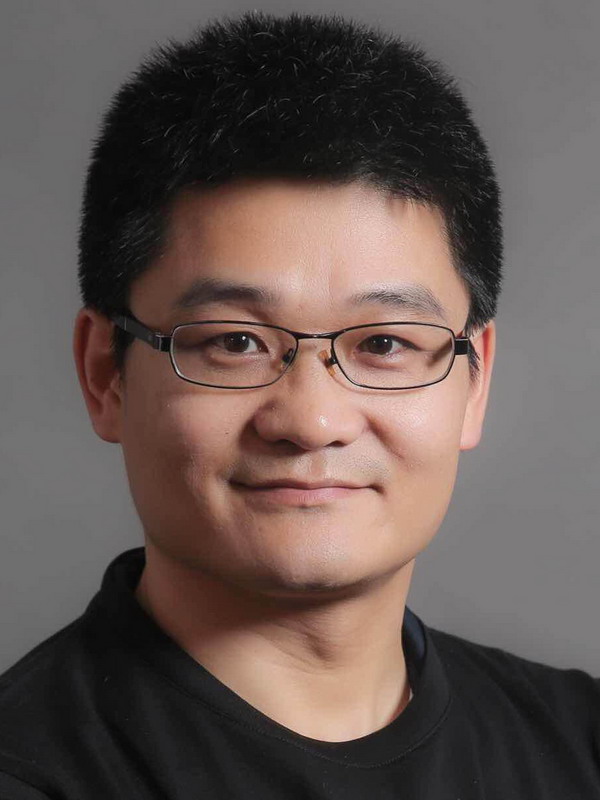}}]{Shiguang Shan}
  received Ph.D. degree in computer science from the Institute of Computing Technology (ICT), Chinese Academy of Sciences (CAS), Beijing, China, in 2004. He has been a full Professor of this institute since 2010 and now the deputy director of CAS Key Lab of Intelligent Information Processing. His research interests cover computer vision, pattern recognition, and machine learning. He has published more than 300 papers, with totally more than 29,000 Google scholar citations. He has served as Area Chair (or Senior PC) for many international conferences including ICCV11, ICPR12/14/20, ACCV12/16/18, FG13/18, ICASSP14, BTAS18, AAAI20/21, IJCAI21, and CVPR19/20/21. And he was/is Associate Editors of several journals including IEEE T-IP, Neurocomputing, CVIU, and PRL. He was a recipient of the China's State Natural Science Award in 2015, and the China's State S\&T Progress Award in 2005 for his research work.
\end{IEEEbiography}





%

\newpage
\appendices
\section{Proof of Proposition~\ref{prop:1}}
\label{sec:prop1}
\begin{proof}
  We use mathematical induction to complete the proof. In the following proof, we use $\bm{g}$ to denote the gradient $g(\bm{x})=\frac{\partial L(\bm{x}) }{\partial \bm{x}}$ and $\bm{H}$ to denote the second-order Hessian matrix $H(\bm{x})$.

  When $m=1$, from \cref{equ:am}-\cref{equ:dm}, we get $a_1=1$, $b_1=0$, $c_1=1$, $d_1=0$. Taking these values into \cref{equ:gm} and \cref{equ:deltam}, we can get $\bm{g_1}=\bm{g}$, $\bm{\delta_1} = \gamma \bm{g}$, which satisfies the formula of MIFGSM~\cite{dong2018boosting} (\ie, \cref{equ:gt} and \cref{equ:deltat}).

  Assuming that the proposition holds when $m=t$, when $m=t+1$,
  \begin{align}
    \bm{g_{t+1}}=&\mu \bm{g_{t}} + g(\bm{x}+\bm{\delta_{t}}) \\
    =&\mu a_t \bm{g} + \mu b_t \gamma \bm{Hg} + \bm{g} + \bm{H\delta_{t}} \\
    =&\mu a_t \bm{g} + \mu b_t \gamma \bm{Hg} + \bm{g} + \bm{H}(c_t \gamma \bm{g}+ d_t \gamma^2 \bm{Hg})   \\
    \approx &(\mu a_t + 1) \bm{g} + (\mu b_t + c_t) \gamma \bm{Hg} 
  \end{align}
  where the second line is due to the first-order Taylor expansion, and we ignore second-order terms with respect to $\bm{H}$ in forth line since $\|\bm{H}\|\sim 0$.
  \begin{align}
    &\mu a_t + 1\nonumber\\
    =&\mu \sum_{i=1}^t \mu^{i-1} + 1\\ 
    =& \sum_{i=1}^t \mu_i + 1\\
    =&\sum_{i=1}^{t+1} \mu_{i-1} = a_{t+1}
  \end{align}
  \begin{align}
    &\mu b_t + c_t \nonumber\\
    =& \mu \sum_{i=1}^t (t-i+1)(i-1) \mu^{i-2} + \sum_{i=1}^t (t-i+1) \mu ^{i-1} \\
    =&\sum_{i=1}^t (t-i+1) (i-1) \mu^{i-1} + \sum_{i=1}^t (t-i+1) \mu ^{i-1} \\
    =&\sum_{i=1}^t (t-i+1) i \mu^{i-1}\\
    =& \sum_{i=2}^{t+1}(t-i+2)(i-1)\mu^{i-2} \\
    =& \sum_{i=1}^{t+1}(t-i+2)(i-1)\mu^{i-2} = b_{t+1}
  \end{align}
  So when $m=t+1$, \cref{equ:gm} holds.
  \begin{align}
    \bm{\delta_{t+1}}=&\bm{\delta_{t}} + \gamma \bm{g_{t+1}} \\
    =&c_t\gamma \bm{g} + d_t \gamma^2 \bm{Hg} + (\mu a_t + 1) \gamma \bm{g} + (\mu b_t + c_t) \gamma^2 \bm{Hg} \\
    =& (\mu a_t + c_t + 1) \gamma \bm{g} + (\mu b_t + c_t+d_t)\gamma^2 \bm{Hg}
  \end{align}
  \begin{align}
     & \mu a_t + c_t + 1\nonumber\\
      =&\mu \sum_{i=1}^t\mu^{i-1}+\sum_{i=1}^t (t-i+1)\mu^{i-1}+1 \\
      =&\sum_{i=2}^{t+1} \mu^{i-1}+\sum_{i=1}^t(t-i+1)\mu^{i-1}+1\\
      =&\sum_{i=2}^t \mu^{i-1} + \sum_{i=2}^t(t-i+1)\mu^{i-1} + \mu^t + t+1\\
      =&\sum_{i=2}^t(t-i+2)\mu^{i-1}+ \mu^t + t+1\\
      =&\sum_{i=1}^{t+1}(t-i+2)\mu^{i-1}=c_{t+1}
    \end{align}
    \begin{align}
      &\mu b_t + c_t + d_t \nonumber\\
      =& \mu \sum_{i=1}^t(t-i+1)(i-1)\mu^{i-2}+\sum_{i=1}^t(t-i+1)\mu^{i-1} \nonumber\\
      &+\sum_{i=1}^t \frac{(t-i+2)(t-i+1)(i-1)}{2}\mu^{i-2} \\
      =& \sum_{i=1}^t(t-i+1)(i-1)\mu^{i-1}+\sum_{i=1}^t(t-i+1)\mu^{i-1} \nonumber\\
      &+\sum_{i=1}^t \frac{(t-i+2)(t-i+1)(i-1)}{2}\mu^{i-2} \\
      =& \sum_{i=2}^{t+1}(t-i+2)(i-2)\mu^{i-2}+\sum_{i=2}^{t+1}(t-i+2)\mu^{i-2}\nonumber\\
      &+\sum_{i=2}^{t+1} \frac{(t-i+2)(t-i+1)(i-1)}{2}\mu^{i-2} \\
      =& \sum_{i=2}^{t+1}(t-i+2)(i-2+1+\frac{(t-i+1)(i-1)}{2})\mu^{i-2}\\
      =& \sum_{i=2}^{t+1}\frac{(t-i+2)(t-i+3)(i-1)}{2}\mu^{i-2} \\
      =& \sum_{i=1}^{t+1}\frac{(t-i+2)(t-i+3)(i-1)}{2}\mu^{i-2}=d_{t+1}
  \end{align}
  So when $m=t+1$, \cref{equ:deltam} holds.

  To sum up, we can get that when $m=t+1$, both \cref{equ:gm} and \cref{equ:deltam} hold. Therefore, Proposition~\ref{prop:1} is proved.
\end{proof}

\section{Proof of Proposition~\ref{prop:2} and Some Discussions}
\label{sec:prop2}
\begin{proof}
  From the Lemma 1 in~\cite{wang2021a}, the Shapley interaction between perturbation units $a, b$ can be written as $I_{ab}=\bm{\delta}_a\bm{H}_{ab}(\bm{x})\bm{\delta}_b +\hat{R_2}(\bm{\delta})$, where $\bm{H}_{ab}(\bm{x})=\frac{\partial L(\bm{x})}{\partial \bm{x}_a \partial \bm{x}_b}$ represents the element of the Hessian matrix, $\bm{\delta}_a$ and $\bm{\delta}_b$ are the $a$-th and $b$-th elements in $\bm{\delta}$, $\hat{R_2}(\bm{\delta})$ denotes terms with elements in $\bm{\delta}$ of higher than the second order. In the following proof, we ignore second-order terms of $\bm{\delta}$ since $\|\bm{\delta}\| \sim 0$. According to Proposition~\ref{prop:1}, the Shapley interaction between perturbation units $a, b$ in our proposed APAA can be further written as
  \begin{align}
    I_{ab}=&\bm{\delta}_a \bm{H}_{ab} \bm{\delta}_b \\
    =& (c_m \gamma \bm{g}_a + d_m \gamma^2  \bm{g}^\top \bm{H}_{*a}) \bm{H}_{ab}(c_m \gamma \bm{g}_b + d_m \gamma^2 \bm{g}^\top\bm{H}_{*b})\\
    =&c_m^2\gamma^2 \bm{g}_a\bm{g}_b\bm{H}_{ab} + c_md_m\gamma^3\bm{g}_b\bm{H}_{ab}\bm{g}^\top \bm{H}_{*a} \nonumber\\
    & +c_md_m\gamma^3\bm{g}_a\bm{H}_{ab}\bm{g}^\top \bm{H}_{*b}+ O(\bm{H}^2),
  \end{align}
  where $O(\bm{H}^2)$ is the second-order small quantity about $\bm{H}$, which is ignored in the following calculations.
  We can further calculate the interaction inside perturbations generated by MIFGSM with APAA as follows
  \begin{align}
    \mathbb{E}_{a,b}(I_{ab})=&\mathbb{E}_{a,b}[c_m^2\gamma^2 \bm{g}_a\bm{g}_b\bm{H}_{ab} + c_md_m\gamma^3\bm{g}_b\bm{H}_{ab}\bm{g}^\top \bm{H}_{*a} \nonumber \\
    &+c_md_m\gamma^3\bm{g}_a\bm{H}_{ab}\bm{g}^\top \bm{H}_{*b}] \\
    =&\gamma^2 \mathbb{E}_{a,b}[c_m^2 \bm{g}_a\bm{g}_b\bm{H}_{ab}] + 2\gamma^3  \mathbb{E}_{a,b}[c_md_m\bm{g}_a\bm{H}_{ab}\bm{g}^\top \bm{H}_{*b}] 
\end{align}
where $\mathbb{E}_{a,b}[c_md_m\bm{g}_a\bm{H}_{ab}\bm{g}^\top \bm{H}_{*b}]$ has been proven greater than 0 in Sec. E.2 in~\cite{wang2021a}. Therefore, the interaction inside adversarial perturbations generated by MIFGSM with APAA at $m$-th step is given as:
\begin{equation}
  \mathbb{E}_{a,b}(I_{ab}) = A\gamma^2+2B\gamma^3,
\end{equation}
where
\begin{equation*}
  A=\mathbb{E}_{a,b}[c_m^2 \bm{g}_a\bm{g}_b\bm{H}_{ab}], \quad B=\mathbb{E}_{a,b}[c_md_m\bm{g}_a\bm{H}_{ab}\bm{g}^\top \bm{H}_{*b}]\geq 0.
\end{equation*}
\end{proof}

\begin{figure}[t]
  \centering
  \includegraphics[width=0.9\columnwidth]{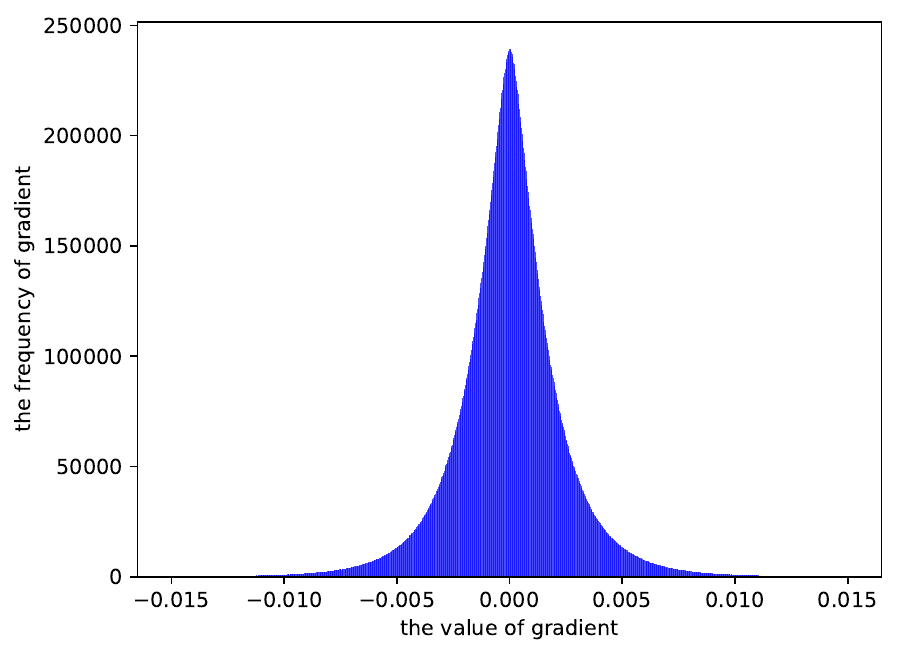}
  \caption{Frequency density histogram of gradients (\ie, $g_t$ in \cref{eqn:g_t}) in the MIFGSM~\cite{dong2018boosting} attack method.}
  \label{fig:gradient}
\end{figure}

When we equivalently transform the sign function in MIFGSM using scaling factor coefficients, in the most ideal scenario, this coefficient may indeed vary with the dimensionality of gradients and the number of attack steps. However, for the sake of convenience and feasibility in our proof, we make a certain simplification by estimating an average, globally consistent scaling factor coefficient to replace the per-dimension-changing coefficients. But we want to emphasize that this simplification does not impact the conclusions of our proof, as we explain from the following two perspectives.
Firstly, we conduct an experiment to visualize the histogram of gradients in MIFGSM method. As shown in~\cref{fig:gradient}, the magnitude of gradients in the MIFGSM method is relatively small, nearly all of them are less than 0.01. If we equivalently convert them to a sign function using scaling factors, it is equivalent to multiplying by a coefficient $\gamma_{MIFGSM}$ greater than $10^2$, which is orders of magnitude larger than the actual scaling factor $\gamma_{APAA}$ used in our APAA method (typically less than 10). In other words, we have $0<\gamma_{APAA} \leq \gamma_{MIFGSM}$. Furthermore, as presented in~\cref{equ:interaction}, the interaction term is a cubic function of $\gamma$, and the coefficient of the cubic term $A$ is greater than 0. Therefore, despite the simplification we made for the MIFGSM method, the significant difference in the magnitudes of equivalent scaling factors between MIFGSM and our method still leads to the conclusion that $\mathbb{E}_{a,b} (I_{ab};\gamma_{APAA})< \mathbb{E}_{a,b} (I_{ab};\gamma_{MIFGSM}$ ).
Secondly, we provide a comparison of histograms of the interaction inside perturbations generated by our APAA and that of MIFGSM in~\cref{fig:interaction} of the manuscript. From the figure, it is clearly that the interaction within perturbations generated by our APAA method is obviously smaller than that of MIFGSM, which is consistent with the conclusion we derived.

\section{More experimental results}
\subsection{More results of comparison between baseline methods and $\text{APAA}_f$ and $\text{APAA}_a$ methods}
To further demonstrate the effectiveness of our proposed APAA method, we presente a comparison between baseline methods and $\text{APAA}_f$ and $\text{APAA}_a$ methods in~\cref{tab:more_cifar_adaptive} across more source models.
As indicated in the table, our proposed APAA method consistently achieves higher attack success rates with a smaller perturbation budget, whether it be on normally trained models or defense models.

\begin{table*}[!hbtp]
  \begin{center}
    \centering
    \caption{More results of the comparisons between the attack success rates of the adversarial examples by baselines, our proposed method with \textbf{fixed scaling factor} (APAA$_f$) and our proposed method with \textbf{adaptive scaling factor} (APAA$_a$) against both normally trained models and defense models on \textbf{CIFAR10} under untargeted attack setting.}
    \label{tab:more_cifar_adaptive}
    \resizebox{\textwidth}{!}{
      \begin{tabular}{c|l|cccccccccc|cc}
      \hline
      \multirow{2}{*}{\begin{tabular}[c]{@{}c@{}}Source\\ Model\end{tabular}} & \multicolumn{1}{c|}{\multirow{2}{*}{Method}} & \multicolumn{10}{c|}{Target Model}  & \multicolumn{2}{c}{Distance Metric} \\ \cline{3-14} 
      & \multicolumn{1}{c|}{}                        & RegNet & Res-18        & SENet-18      & DPN  & \multicolumn{1}{c|}{ShakeShake} & Dense-121\textsubscript{adv}    & GoogLeNet\textsubscript{adv} & Res-18\textsubscript{adv} & $k$-WTA & Ensemble & MAD              & RMSD              \\ \hline
      \multirow{12}{*}{\begin{tabular}[c]{@{}c@{}}Res-18\\+$k$-WTA\end{tabular}}
      & MIFGSM~\cite{dong2018boosting}    & 86.4 & 99.4 & 93.5 & 82.7 & \multicolumn{1}{c|}{89.2} & 16.0 & 37.4 & 15.8 & 90.6 & 84.0 & 5.520 & 5.873 \\
      & MIFGSM w/ APAA$_f$                 & 90.5 & 99.9 & 96.9 & 86.0 & \multicolumn{1}{c|}{93.4} & 19.8 & 44.3 & 19.2 & 96.0 & 88.4 & \textbf{5.408} & \textbf{5.765} \\   
      & MIFGSM w/ APAA$_a$                 & \textbf{90.9} & \textbf{100.0} & \textbf{97.4} & \textbf{86.6} & \multicolumn{1}{c|}{\textbf{93.9}} & \textbf{22.9} & \textbf{48.8} & \textbf{19.4} & \textbf{96.9} & \textbf{89.0} & 5.763 & 6.375 \\   \cline{2-14}
      & DIM~\cite{xie2019improving}        & 89.9 & 99.3 & 95.3 & 86.5 & \multicolumn{1}{c|}{92.0} & 17.8 & 43.9 & 17.9 & 90.6 & 87.6 & 5.576 & 5.910 \\
      & DIM w/ APAA$_f$                    & 93.3 & 99.8 & 97.8 & 89.8 & \multicolumn{1}{c|}{95.3} & 23.0 & 51.6 & 20.7 & 94.8 & 91.5 & 5.323 & 5.679 \\
      & DIM w/ APAA$_a$                    & \textbf{94.1} & \textbf{99.9} & \textbf{98.3} & \textbf{90.4} & \multicolumn{1}{c|}{\textbf{96.1}} & \textbf{23.5} & \textbf{54.0} & \textbf{21.9} & \textbf{95.9} & \textbf{92.4} & \textbf{5.063} & \textbf{6.324} \\   \cline{2-14}
      & SIM~\cite{lin2020nesterov}         & 91.7 & 99.4 & 96.6 & 88.2 & \multicolumn{1}{c|}{93.6} & 19.6 & 47.2 & 19.0 & 92.0 & 90.5 & 5.620 & 5.948 \\
      & SIM w/ APAA$_f$                    & 94.8 & \textbf{99.9} & 98.6 & 91.4 & \multicolumn{1}{c|}{96.6} & 24.6 & 55.3 & 21.9 & 95.7 & 93.3 & \textbf{5.339} & \textbf{5.689} \\
      & SIM w/ APAA$_a$                    & \textbf{95.4} & \textbf{99.9} & \textbf{98.8} & \textbf{91.6} & \multicolumn{1}{c|}{\textbf{97.1}} & \textbf{26.1} & \textbf{58.0} & \textbf{22.7} & \textbf{96.4} & \textbf{93.9} & 5.712 & 6.324 \\   \cline{2-14}
      & SVRE~\cite{xiong2022stochastic}    & 94.6 & 99.9 & 98.8 & 91.8 & \multicolumn{1}{c|}{97.5} & 23.4 & 57.9 & 17.4 & 97.7 & 93.9 & 5.770 & 6.097 \\
      & SVRE w/ APAA$_f$                   & 96.6 & \textbf{100.0} & 99.2 & 93.7 & \multicolumn{1}{c|}{98.1} & 26.8 & 60.8 & 20.1 & 98.5 & 95.6 & 5.394 & 5.753 \\
      & SVRE w/ APAA$_a$                   & \textbf{96.9} & \textbf{100.0} & \textbf{99.5} & \textbf{94.0} & \multicolumn{1}{c|}{\textbf{98.6}} & \textbf{27.9} & \textbf{61.7} & \textbf{21.8} & \textbf{98.9} & \textbf{96.1} & \textbf{5.131} & \textbf{5.498} \\    \hline
      \multirow{12}{*}{\begin{tabular}[c]{@{}c@{}}Res-18\\+GoogLeNet\textsubscript{adv}\end{tabular}}
      & MIFGSM~\cite{dong2018boosting}    & 89.5 & 96.5 & 93.1 & 86.2 & \multicolumn{1}{c|}{90.2} & 45.0 & 100.0 & 24.4 & 83.4 & 86.7 & 5.180 & 5.618 \\
      & MIFGSM w/ APAA$_f$                 & 93.0 & 98.5 & 95.8 & 89.5 & \multicolumn{1}{c|}{93.8} & \textbf{53.8} & \textbf{100.0} & \textbf{30.1} & 87.0 & 89.8 & \textbf{4.989} & \textbf{5.413} \\
      & MIFGSM w/ APAA$_a$                 & \textbf{93.2} & \textbf{98.9} & \textbf{96.1} & \textbf{89.8} & \multicolumn{1}{c|}{\textbf{93.9}} & 53.1 & \textbf{100.0} & 29.9 & \textbf{87.2} & \textbf{90.1} & 5.049 & 5.789 \\   \cline{2-14}
      & DIM~\cite{xie2019improving}        & 90.4 & 96.2 & 93.3 & 87.3 & \multicolumn{1}{c|}{90.8} & 40.7 & 99.2 & 24.6 & 85.3 & 87.9 & 5.345 & 5.751 \\
      & DIM w/ APAA$_f$                    & 93.3 & 98.2 & 95.8 & 90.3 & \multicolumn{1}{c|}{93.6} & \textbf{47.4} & 99.8 & 29.7 & 88.2 & 90.9 & \textbf{4.994} & \textbf{5.405} \\
      & DIM w/ APAA$_a$                    & \textbf{93.5} & \textbf{98.6} & \textbf{96.4} & \textbf{90.4} & \multicolumn{1}{c|}{\textbf{93.9}} & 47.3 & \textbf{99.9} & \textbf{29.8} & \textbf{88.7} & \textbf{91.0} & 5.024 & 5.763 \\   \cline{2-14}
      & SIM~\cite{lin2020nesterov}         & 95.4 & 99.0 & 97.3 & 92.8 & \multicolumn{1}{c|}{95.5} & 49.4 & 97.5 & 30.1 & 91.8 & 94.1 & 5.789 & 6.401 \\
      & SIM w/ APAA$_f$                    & 96.7 & 99.5 & 98.5 & 94.5 & \multicolumn{1}{c|}{96.9} & \textbf{54.7} & 98.7 & 33.6 & 93.0 & 95.3 & 5.158 & 5.893 \\
      & SIM w/ APAA$_a$                    & \textbf{96.9} & \textbf{99.6} & \textbf{98.7} & \textbf{94.8} & \multicolumn{1}{c|}{\textbf{97.3}} & \textbf{54.7} & \textbf{99.0} & \textbf{33.7} & \textbf{93.4} & \textbf{95.4} & \textbf{5.151} & \textbf{5.891} \\   \cline{2-14}
      & SVRE~\cite{xiong2022stochastic}    & 96.0 &	99.5 & 98.1 &	94.0 & \multicolumn{1}{c|}{96.4} & 51.3 & 98.7 & 28.6 & 92.5 & 94.8 & 5.617 & 5.998 \\
      & SVRE w/ APAA$_f$                    & 97.5 & 99.8 & 99.0 & 95.3 & \multicolumn{1}{c|}{97.7} & 55.7 & 99.6 & 29.4 & \textbf{94.1} & 96.0 & 5.197 & 5.594 \\
      & SVRE w/ APAA$_a$                    & \textbf{97.9} & \textbf{99.9} & \textbf{99.3} & \textbf{95.7} & \multicolumn{1}{c|}{\textbf{98.0}} & \textbf{55.9} & \textbf{99.8} & \textbf{30.1} & 94.0 & \textbf{96.3} & \textbf{5.130} & \textbf{5.413} \\    \hline
      \end{tabular}
    }
  \end{center}
\end{table*}

\subsection{Comparison under the setting of $L_2$ attack}
To further elucidate the generalizability of our proposed APAA method, We conduct a comparison between our $\text{APAA}_a$ and baseline methods under the $L_2$ attack setting. When applied to $L_2$ attack, our $\text{APAA}_a$ can be simply reformulated as $\bm{x^{adv}_{t+1}} = \Pi_{\bm{x}, \epsilon} (\bm{x_t^{adv}} + \gamma_{t+1} \cdot \frac{\bm{g_{t+1}}}{\|\bm{g_{t+1}}\|_2}),$ where $\bm{x_t^{adv}}$ represents the adversarial example generated in the $t$-th step, $\Pi_{\bm{x},\epsilon}$ means to clip the generated adversarial examples within the $\epsilon$-neighborhood of the original image on $L_2$ norm, $\gamma_t$ represents the scaling factor generated by our $\text{APAA}_a$ in the $t$-th step, $\bm{g_t}$ is the accumulated gradient defined in~\cref{eqn:g}. In contrast, $\gamma_t$ is fixed step size in the baseline methods. As shown in~\cref{tab:cifar_l2}, it is obvious that under the $L_2$ attack setting, our proposed $\text{APAA}_a$ consistently achieves higher attack success rates than all the compared baseline methods, while with smaller adversarial perturbation budgets. This effectively demonstrates the well effectiveness and generalizability of our proposed approach.
\begin{table*}[!t]
  \begin{center}
    \centering
    \caption{The attack success rates of the adversarial examples on \textbf{CIFAR10} under \textbf{$L_2$ attack} setting.}
    \label{tab:cifar_l2}
    \begin{subtable}[t]{\textwidth}
      \caption{The evaluation on the normally trained models.}
      \resizebox{\textwidth}{!}{
        \begin{tabular}{c|l|cccccccc|cc}
        \hline
        \multirow{2}{*}{\begin{tabular}[c]{@{}c@{}}Source\\ Model\end{tabular}} & \multicolumn{1}{c|}{\multirow{2}{*}{Method}} & \multicolumn{8}{c|}{Target Model}                                                          & \multicolumn{2}{c}{Distance Metric} \\ \cline{3-12} 
        & \multicolumn{1}{c|}{}                        & RegNet        & Res-18        & SENet-18       & Dense-121     & WideRes\textsubscript{28$\times$10}    & DPN    & Pyramid    & ShakeShake    & MAD                  & RMSD                 \\ \hline
        \multirow{6}{*}{\begin{tabular}[c]{@{}c@{}}Res-18\\+SENet-18\\+Dense-121\textsubscript{adv}\\+GoogLeNet\textsubscript{adv}\end{tabular}}        & MIFGSM~\cite{dong2018boosting}              & 97.3 & 99.0 & 99.7 & 98.3 & 98.1 & 96.4 & 97.1 & 97.5 & 5.636 & 7.982 \\
        & MIFGSM w/ APAA$_a$            & \textbf{99.0} & \textbf{99.8} & \textbf{99.9} & \textbf{99.6} & \textbf{99.4} & \textbf{98.2} & \textbf{98.1} & \textbf{99.2} & \textbf{4.817} & \textbf{5.459} \\  \cline{2-12} 
        & DIM~\cite{xie2019improving}   & 97.7 & 98.9 & 99.4 & 98.3 & 98.1 & 96.6 & 97.3 & 97.8 & 5.610 & 7.982 \\
        & DIM w/ APAA$_a$               & \textbf{99.1} & \textbf{99.9} & \textbf{99.9} & \textbf{99.6} & \textbf{99.4} & \textbf{98.4} & \textbf{98.2} & \textbf{99.4} & \textbf{4.801} & \textbf{5.437} \\  \cline{2-12} 
        & SIM~\cite{lin2020nesterov}    & 98.8 & 99.5 & 99.8 & 99.3 & 99.2 & 98.1 & 98.8 & 99.0 & 5.610 & 7.982 \\
        & SIM w/ APAA$_a$               & \textbf{99.8} & \textbf{100.0} & \textbf{100.0} & \textbf{99.9} & \textbf{99.8} & \textbf{99.5} & \textbf{99.6} & \textbf{99.8} & \textbf{4.844} & \textbf{5.453} \\      \hline
        \end{tabular}
      }
    \end{subtable}
    \begin{subtable}[t]{0.9\textwidth}
      \caption{The evaluation on the defense models.}
      \resizebox{\textwidth}{!}{
        \begin{tabular}{c|l|ccccccc}
        \hline
        \multirow{2}{*}{\begin{tabular}[c]{@{}c@{}}Source\\ Model\end{tabular}} & \multicolumn{1}{c|}{\multirow{2}{*}{Method}} & \multicolumn{7}{c}{Target Model}              \\ \cline{3-9} 
        & \multicolumn{1}{c|}{}                        & Dense-121\textsubscript{adv}   & GoogLeNet\textsubscript{adv}   & Res-18\textsubscript{adv}   & $k$-WTA   & Odds   & Generative   & Ensemble   \\ \hline
        \multirow{6}{*}{\begin{tabular}[c]{@{}c@{}}Res-18\\+SENet-18\\+Dense-121\textsubscript{adv}\\+GoogLeNet\textsubscript{adv}\end{tabular}}       & MIFGSM~\cite{dong2018boosting}              & 98.6 & 99.9 & 99.4 & 96.0 & 99.1 & 85.2 & 96.6 \\
        & MIFGSM w/ APAA$_a$            & \textbf{100.0} & \textbf{100.0} & \textbf{99.9} & \textbf{98.1} & \textbf{99.6} & \textbf{86.4} & \textbf{98.1} \\  \cline{2-9} 
        & DIM~\cite{xie2019improving}   & 97.6 & 99.7 & 99.6 & 96.2 & 99.0 & 85.5 & 96.8 \\
        & DIM w/ APAA$_a$               & \textbf{99.9} & \textbf{100.0} & \textbf{99.7} & \textbf{98.4} & \textbf{99.7} & \textbf{86.2} & \textbf{98.4} \\   \cline{2-9} 
        & SIM~\cite{lin2020nesterov}    & 98.0 & 99.1 & 99.1 & 98.0 & 99.5 & 89.2 & 98.3 \\
        & SIM w/ APAA$_a$               & \textbf{99.8} & \textbf{99.9} & \textbf{99.5} & \textbf{99.2} & \textbf{100.0} & \textbf{89.8} & \textbf{99.3} \\   \hline
        \end{tabular}
      }
    \end{subtable}
  \end{center}
\end{table*}

\subsection{Distribution of scaling factors at different attack steps}
\begin{figure*}[t]
  \centering
  \begin{subfigure}[t]{0.4\textwidth}
    \centering
    \includegraphics[width=\textwidth]{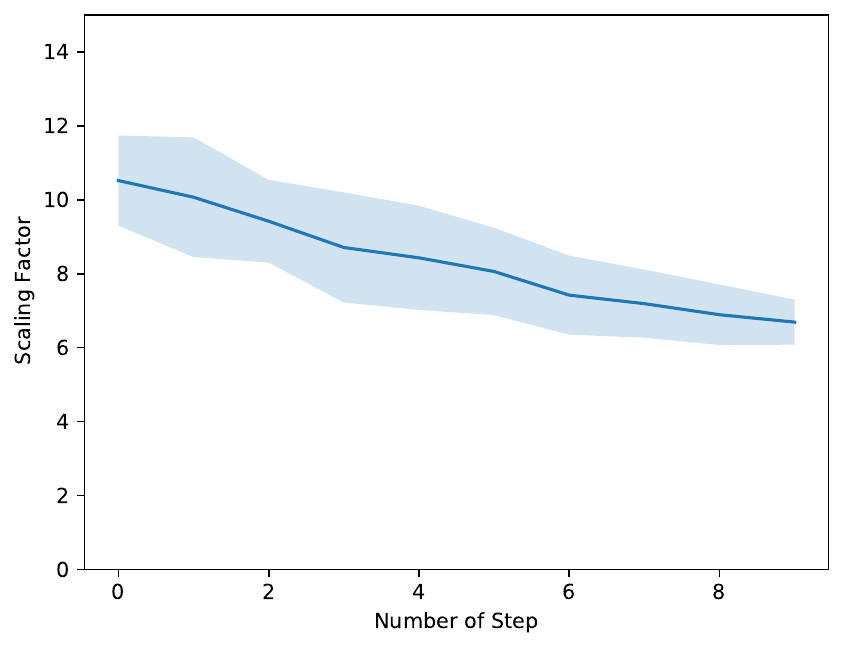}
    \caption{Res-18+GoogLeNet\textsubscript{adv}}
    \end{subfigure}
  \quad
  \begin{subfigure}[t]{0.4\textwidth}
    \centering
    \includegraphics[width=\textwidth]{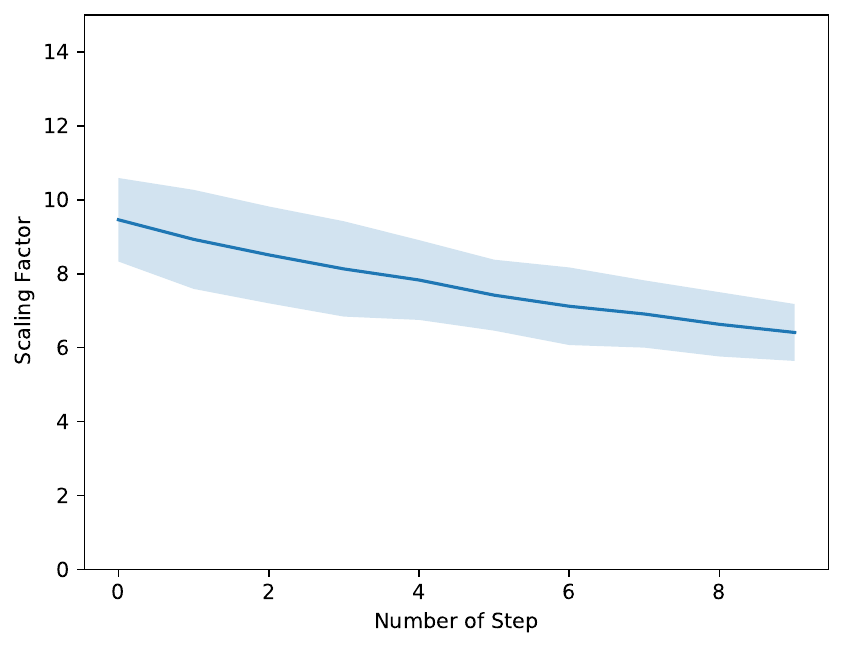}
    \caption{Res-18+SENet-18+Dense-121\textsubscript{adv}+GoogLeNet\textsubscript{adv}}
  \end{subfigure}
  \caption{The statistics of the scaling factors generated at each step in $\text{APAA}_a$. The shaded part represents the standard deviation corresponding to the scaling factor of each step. The experiment is conducted based on MIFGSM~\cite{dong2018boosting} method. The caption in each sub-figure indicates the source model used for training.}
  \label{fig:scaling}
\end{figure*}

In order to gain a deeper insight into the scaling factor generator we have trained, we provide the statistics of the learned scaling factors under different attack steps in~\cref{fig:scaling}. As shown in the figure, as aligned with our earlier hypothesis that larger scaling factors can be employed in the initial steps to swiftly approach the optima, followed by a gradual reduction in step size to finetune the adversarial perturbation. This visualization enhances the comprehension of our approach's adaptability throughout the attack process.

\end{document}